\documentclass[twoside,11pt]{article}
\usepackage[preprint]{jmlr2e}
\usepackage{graphicx}
\usepackage{amsmath}
\usepackage{amsfonts}
\usepackage{multirow, makecell}
\usepackage{mathtools}
\usepackage{cancel}
\usepackage{booktabs,threeparttable}
\usepackage{float}
\usepackage[caption=false, font=footnotesize]{subfig}
\usepackage[shortlabels]{enumitem}
\usepackage[noend]{algorithm2e}
\usepackage{textcomp}
\usepackage{etoolbox}
\usepackage{diagbox}
\usepackage{microtype}
\usepackage{anyfontsize}
\usepackage{xfrac}

\newtheorem{assumption}[theorem]{Assumption}

\usepackage{lastpage}
\jmlrheading{XX}{2024}{1-\pageref{LastPage}}{12/24; Revised XX}{XX}{21-0000}{Saurab Chhachhi and Fei Teng}

\ShortHeadings{Wasserstein Markets for Differentially-Private Data}{Chhachhi and Teng}
\firstpageno{1}

\begin{document}

\title{Wasserstein Markets for Differentially-Private Data}

\author{\name Saurab Chhachhi \email saurab.chhachhi11@imperial.ac.uk \\
       \addr Department of Electrical \& Electronic Engineering\\
       Imperial College London\\
       London, SW7 2AZ, United Kingdom
       \AND
       \name Fei Teng \email f.teng@imperial.ac.uk \\
       \addr Department of Electrical \& Electronic Engineering\\
       Imperial College London\\
       London, SW7 2AZ, United Kingdom}

\editor{My editor}

\maketitle
\begin{abstract}
    Data is an increasingly vital component of decision making processes across industries. However, data access raises privacy concerns motivating the need for privacy-preserving techniques such as differential privacy. Data markets provide a means to enable wider access as well as determine the appropriate privacy-utility trade-off. Existing data market frameworks either require a trusted third party to perform computationally expensive valuations or are unable to capture the combinatorial nature of data value and do not endogenously model the effect of differential privacy. This paper addresses these shortcomings by proposing a valuation mechanism based on the Wasserstein distance for differentially-private data, and corresponding procurement mechanisms by leveraging incentive mechanism design theory, for task-agnostic data procurement, and task-specific procurement co-optimisation. The mechanisms are reformulated into tractable mixed-integer second-order cone programs, which are validated with numerical studies.
\end{abstract}


\section{Introduction}
Machine learning is being rapidly adopted by a range of industries as they recognise the value of data-driven decision making and analytics \citep{Agarwal2019}. This is contingent on the availability of large amounts of high quality data. Existing practices assume the decision-maker or data user has free access to the required data streams, which often consists of personal information, and therefore accrues all the benefits of data access. However, this does not account for the data costs such as privacy concerns \citep{Veliz2018} and commercial sensitivity \citep{Goncalves2020}. As a result, a growing body of literature has been investigating the use of Privacy-Preserving Techniques (PPT) \citep{Rubaie2019} and data markets to enable wider access to data \citep{Bergemann2019}.

Although a wide range of PPTs have been proposed as a means to balance privacy and data access, Differential Privacy (DP), has gained significant traction given the strong formal privacy guarantees it provides \citep{Teng2022, Thomas2021}. However, DP, which enables access to aggregated data through noise addition, while maintaining individual data owners' privacy, introduces an inevitable Privacy-Utility Trade-off (PUT) \citep{Chhachhi2021}. This can impact the optimality of decision making \citep{Elroy2023} and hence the value of the data. Data markets incentivise data owners to share their data by compensating them for the value that their data provides \citep{Agarwal2019} while also providing a means to determine the appropriate PUT.

Data markets can broadly be defined by two components: a valuation scheme, i.e., how data value is quantified, and a procurement mechanism, i.e., how payments for data sharing are determined. Existing proposals for data markets either employ a cooperative game (CG) approach \citep{Agarwal2019,Liu2021,Goncalves2020,Han2020,Pinson2021}, or an incentive mechanism design (IMD) approach \citep{Ren2022,Zhang2021,Zhao2018a,Jiao2019} as their procurement mechanism with existing data valuation schemes being broadly applicable to either.

CGs assume that by sharing data and entering into a coalition, value (e.g., model accuracy) will be improved. The coalition with all participants, the grand coalition, is usually assumed to be the coalition with maximum value. The aim of the market platform is to determine a payment policy which ensures incentive compatibility (IC) i.e., that each participant is no worse off in the grand coalition than in another subset, and individual rationality (IR) i.e., each participant is no worse off by participating in the game. Most CG structures use the marginal improvements in a particular performance metric for a specific task as a valuation metric and then use the Shapley Value or other semi-values as the basis for data payments \citep{Agarwal2019, Lin2024}. For example, \cite{Pinson2021} uses the reduction in mean squared error (MSE) of an hour-ahead wind forecast and \cite{Han2020} uses the improvement in a retailer's energy procurement profits. As such, CGs are able to capture the combinatorial nature of data value i.e., it's dependence on other available data. However, extant literature employing this approach require a Trusted Third Party (TTP), the market platform, to access both the data and model under consideration, and the resulting mechanisms are computationally intensive, rising exponentially with the number of data owners \citep{Jia2019}. Data is a unique commodity that can be reused for multiple purposes at zero marginal cost \citep{Agarwal2019}. Therefore, valuation based on a single task may not be reflective of the potential value derived from the data. Furthermore, CG approaches are vulnerable to manipulation through model mis-specification. In addition, existing CG approaches also do not allow for the modelling of PUT explicitly as they do not endogenously model the effect of DP on data value. Although the framework in \cite{Liu2021} allows data owners to specify their privacy preferences in terms of DP, these are not explicitly linked to data value. Finally, CGs inherently assume data owners will share their data and thus do not have privacy concerns or corresponding reserve prices \citep{Han2020}. 

In contrast, IMD approaches assume data and data owners' reserve prices are held privately and only a data valuation metric is shared with the data market. The platform aims to ensure data owners report their reserve prices truthfully (IC) and they are paid at least this price if their data is purchased (IR). A variety of data valuation metrics have been proposed including statistical distances, (e.g., Jensen-Shannon Divergence (JSD) in \cite{Ren2022}, Wasserstein Distance (WD) in \cite{Jiao2019,Jahani2024}, and Kullback-Leibler Divergence (KLD) in \cite{Falconer2024}), the generalisation error of a Distributionally-Robust Optimisation (DRO)\citep{Lin2024}, as well as the DP privacy budget, $\epsilon$\citep{Zhang2021}. Statistical distances, measures of the difference between a data distribution and some reference distribution, provide a task-agnostic notion of value and are therefore representative of value across use cases. However, this inherently results in the connection between the data valuation metric and performance being lost. An exception is \cite{Falconer2024} which proposes the KLD between the output predictive distribution with and without the data features for a Bayesian regression task. However, this again requires a TTP. To overcome these issues \cite{Zhao2018a} bound the loss of a federating learning algorithm by bounding the weight divergence using the WD. \cite{Jiao2019} instead simulates the relationship to estimate parameters for a pre-specified transfer function, which could be computationally expensive depending on the given task. The use of statistical distances also raises issues around how value should be consolidated across individuals. For example, \citep{Ren2022,Zhang2021} assume value is additive, and  \citep{Jiao2019,Zhao2018a} calculate a weighted average of individual distances. However, neither approach is theoretically grounded nor do they capture the combinatorial nature of data value.

\cite{Ren2022} uses DP to ensure the valuation process is privacy-preserving, removing the need for a TTP to operate the data market, but does not consider the implications of DP noise addition on data value. On the other hand, \citep{Zhang2021} explicitly model the effect of DP on data value but consider an I.I.D. setting where data is only differentiated by data owners' privacy preferences. The IMD approaches described above, all assume an exogenous budget is provided by the data buyer. However, in many applications the available budget for data procurement depends on the value that data provides. This decision-dependent structure is modelled in \cite{Fallah2022}, but only for a specific task, namely mean estimation with I.I.D. differentially-private data, where data owners have heterogeneous privacy preferences. \cite{Pandey2023} consider a more general class of regression markets but the effect of DP is modelled using a pre-specified transfer function. \cite{Mieth2023} provides an alternative approach using DRO with a WD ambiguity set. This is able to model the effect of DP on value explicitly through the WD but still requires data to be shared with the buyer/market platform prior to procurement and does not incorporate data owners' reserve prices. Finally, \cite{Jahani2024} proposes a similar method for privacy-preserving data valuation using the 2-WD but also requires that the distributions under considerations be parametrised as Gaussians. 

In this paper we propose a novel data market framework which addresses the shortcomings outlined above. Specifically, we make the following contributions:
\begin{itemize}
    \item We propose a data valuation mechanism for aggregated differentially-private data based on the WD. It provides a task-agnostic valuation metric which endogenously models the effect of DP. We show that its calculation is privacy-preserving, forgoing the need to share datasets with the buyer or market platform prior to valuation or procurement.
    \item We develop three novel procurement mechanisms, based on IMD, which leverage the proposed data valuation mechanism: 1) an exogenous budget feasible mechanism which incorporates both a non-I.I.D. setting and endogenously captures the effect of DP for task-agnostic procurement, 2) an endogenous budget mechanism, 3) a joint optimisation mechanism. The latter two use Lipschitz bounds to capture the decision-dependence of data procurement for task-specific procurement.
    \item We provide a solution method for the proposed mechanisms using a Hoeffding bound approximation and reformulation as a tractable Mixed-Integer Second-Order Cone Program (MISOCP). The performance of the proposed mechanisms is validated and the trade-offs introduced by the approximation bounds are explored with extensive simulation studies using synthetic data.
\end{itemize}

The remainder of the paper is organised as follows. Section \ref{sec:data_val} presents the data valuation framework outlining the suitability of the WD as a data value metric. The procurement mechanisms and reformulations are introduced in Section \ref{sec:procure}. Numerical case studies for parameter estimation using synthetic data are provided in Section \ref{sec:case}. Finally, conclusions are drawn and future research directions discussed in Section \ref{sec:conc}.

\section{Data Valuation Framework} \label{sec:data_val}
We consider a setting where a data buyer is aiming to purchase data from a set of $N$ data owners. Each data owner has a private dataset, $X_i$, where $i \in \mathcal{N}$\footnote{Although we focus on data procurement, our framework could be extended to value model updates for federated learning \citep{Zhao2018a} or information/forecasts under prediction markets \citep{Storkey2011}.}. The true/target data distribution $X_T$ is an aggregation of all data owners' data, $\mathcal{X}$. In this paper we restrict ourselves to the Euclidean aggregate, $X_T = \frac{1}{N}\sum_{i \in \mathcal{N}} X_i$, however the framework could be adapted to other aggregations such as the Wasserstein barycenter considered in the forecast trading mechanism in \cite{Raja2023}. The dataset obtained by the data buyer is privacy-preserving using local DP as described in \cite{Fallah2022}\footnote{We limit ourselves to this DP formulation given the issues, such as sequential composition, associated with other task/application specific formulations \citep{Blanco2023}.}. The dataset received by the data buyer, $X_P$ is therefore an aggregation of the procured datasets from a subset $P \subseteq \mathcal{N}$ of data owners, where each procured dataset has been locally perturbed using either the Laplace or Gaussian mechanism \citep{Dwork2014}. The data buyer is interested in procuring a subset of data which best represents, $X_T$, the true distribution. Importantly, this objective is not necessarily linked to the specific task/set of tasks the data buyer may wish to use the data for. This naturally motivates the use of statistical distances. The buyer and the data market platform are not assumed to be trusted, therefore the valuation and procurement mechanisms must also be privacy-preserving.

This section first motivates the use of the WD as an appropriate statistical distance, and then proceeds to develop the proposed analytical framework for data valuation. This includes translating the WD into task-specific performance guarantees, endogenously modelling the effect of DP and an efficient approximation scheme for the combinatorial nature of data value. Finally we also briefly discuss the private computation of the WD.

\subsection{Wasserstein Distance as a Valuation Metric} \label{sec:wd_motiv}  There are a wide range of statistical distances and divergences with different properties, providing insights along different dimensions of probability distributions. A comprehensive review, including the relationship between different distances, can be found in \citep{Gibbs2002}. Here, we focus on five popular distances/divergences which have been proposed in the context of data valuation: the Kullback-Leibler Divergence (KLD), Total Variational Distance (TVD), Kolmogorov-Smirnov Metric (KS), JSD, and WD (specifically the 1-Wasserstein Distance). 

To compare the statistical distances above, we set out a number of desirable properties for their use as data valuation metrics. First, whether the distance is a true metric and therefore obeys the associated four axioms\citep{Panaretos2019}: (1) identity of indiscernibles $d(X_1,X_1)=0$, (2) symmetry $d(X_1,X_2) = d(X_2,X_1)$, (3) triangle inequality $d(X_1,X_3) \leq d(X_1,X_2) + d(X_2,X_3)$, and  (4) non-negativity $d(X_1,X_2) \geq 0$. As we are considering relative differences in performance, $L(X_1)-L(X_2) = - \left(L(X_2)-L(X_1)\right)$, it is desirable for the distance to be symmetric and equal to zero when $X_1=X_2$. In addition, as we are motivated by the combined value of multiple data, additivity or in this case sub-additivity (triangle inequality) is a useful feature. Indeed, we use this and the non-negativity property to develop approximation bounds in Section  \ref{sec:metric_dp} and \ref{sec:metric_hoeffding}. The next desirable property is whether the distance is non-saturating and therefore provides meaningful values across inputs. Saturation may be seen as a useful property as it can be used to model the law of diminishing returns, a common assumption in data valuation\citep{Chen2021}. However, we argue that it is restrictive for the valuation metric itself to exhibit these dynamics and should instead be modelled explicitly as a function of data quantity not data quality.

\begin{table}[ht]
    \centering
    \caption{Properties of Statistical Distance/Divergences}
    \begin{threeparttable}
        \begin{tabular}{ccccc}
            \toprule
            \multirow{2}{*}{\makecell{Measure}} &  \multirow{2}{*}{\makecell{Metric}} & \multirow{2}{*}{\makecell{Non-\\Saturating}} & \multirow{2}{*}{\makecell{Disjoint \\ Supports}} & \multirow{2}{*}{\makecell{Input}}\\
            &&&&\\
            \midrule
            KLD & & \checkmark & & $f_{X_1}, f_{X_2}$\\
            JSD & \checkmark  &  & \checkmark & \makecell{$f_{X_1}, f_{X_2},$ \\$f_{\frac{X_1+X_2}{2}}$}\\
            KS & \checkmark & &\checkmark & $F_{X_1}, F_{X_2}$\\ 
            WD & \checkmark & \checkmark & \checkmark & $F_{X_1}, F_{X_2}$\\
            TVD & \checkmark &  & \checkmark & $f_{X_1}, f_{X_2}$\\
            \bottomrule
        \end{tabular}
        \begin{tablenotes}
            \item $f_{X_i}$ and $F_{X_i}$ denote the PDF and CDF, respectively.
        \end{tablenotes}
        \end{threeparttable}
    \label{tab:stat_distances}
\end{table}

Another important consideration is whether the distance is defined when the two distribution under considerations have disjoint or non-overlapping supports, which is especially relevant for empirical data. Finally, statistical distances can be defined either in terms of the distributions' the cumulative distribution functions (CDF) or probability density functions (PDF). When working with empirical data, distances defined based on CDFs are more attractive as they avoid the need to estimate PDFs, either using distributional assumptions on the data or using distribution-free methods such as kernel density estimation, which can be computationally intensive and prone to significant error for smaller datasets.

The criteria are summarised in Table \ref{tab:stat_distances}. The KLD, is non-saturating but does not meet any of the other criteria namely, it is not a metric as it is not symmetric and does not obey the triangle inequality, it requires PDFs to calculate and is infinite when the supports of the distributions being compared are not the same. The JSD, a symmetrisation of the KLD, overcomes some of these issues as it is a metric and is defined for disjoint supports. However, the JSD still requires the calculation of PDFs as well as a mid-point distribution, $f_{\frac{X_1+X_2}{2}}$. In addition, the JSD is bounded and thus exhibits saturating behaviour. The TVD has similar limitations. The WD and KS rely on CDFs, are defined for disjoint supports and are metrics, however, the KS is bounded and saturating. Therefore, the WD exhibits all the desired characteristics while also taking into account the metric space i.e., the actual distance between points in the two distributions rather than their distance in probability.

\subsection{Performance Guarantees - Lipschitz Bound}
The WD has been used in a range of applications such as generative adversarial networks \citep{Arjovsky2017}, distributionally robust optimisation \citep{Esfahani2018}, bounding generalisation errors of machine learning models \citep{Lopez2018}, and recently as the loss function for probabilistic forecasting of wind power \citep{Hosseini2023}. Interestingly, the WD also provides a natural way to link the input and output space. The dual formulation of the WD provides an alternative interpretation as the error in the expected value of 1-Lipschitz functions, $h$, due to the approximation of one distribution by another \citep{Panaretos2019}: 
\begin{align}
    d_{W}(X_1,X_2) = \sup_{\left\lVert h \right\rVert_{Lip} \leq 1} \left\lvert{ \int h dX_1 - \int h dX_2 }\right\rvert
    \label{eq:wass_dual}
\end{align}

The WD provides a task-agnostic measure of data value, however, in many applications data procurement is linked to a particular task and/or model, for example, electricity load forecasting. As such, it is desirable to relate the WD between a distribution, $X_P$, and the target distribution, $X_T$, to the performance difference, $L_{\mathcal{M}}(X_P) - L_{\mathcal{M}}(X_T)$, achieved using the two distributions, for a specific task $\mathcal{M}$ and associated metric, $L_{\mathcal{M}}(X)= \mathbb{E}[l_{\mathcal{M}}(x)]$. This differentiates our proposed framework from existing data valuation mechanisms that use the WD. Specifically, we aim to provide a generic framework which provides performance guarantees for a wide range of (potentially stacked) tasks/models allowing for both task-specific and task-agnostic procurement. In contrast, the IMD approach in \cite{Jiao2019} requires the calculation of pre-specified transfer functions, \cite{Zhao2018a} is limited to federated learning applications, \cite{Jahani2024} is completely task-agnostic, and the DRO approach in \cite{Mieth2023} is task-specific.

\begin{figure}[htb]
    \centering
    \includegraphics[width=0.6\textwidth]{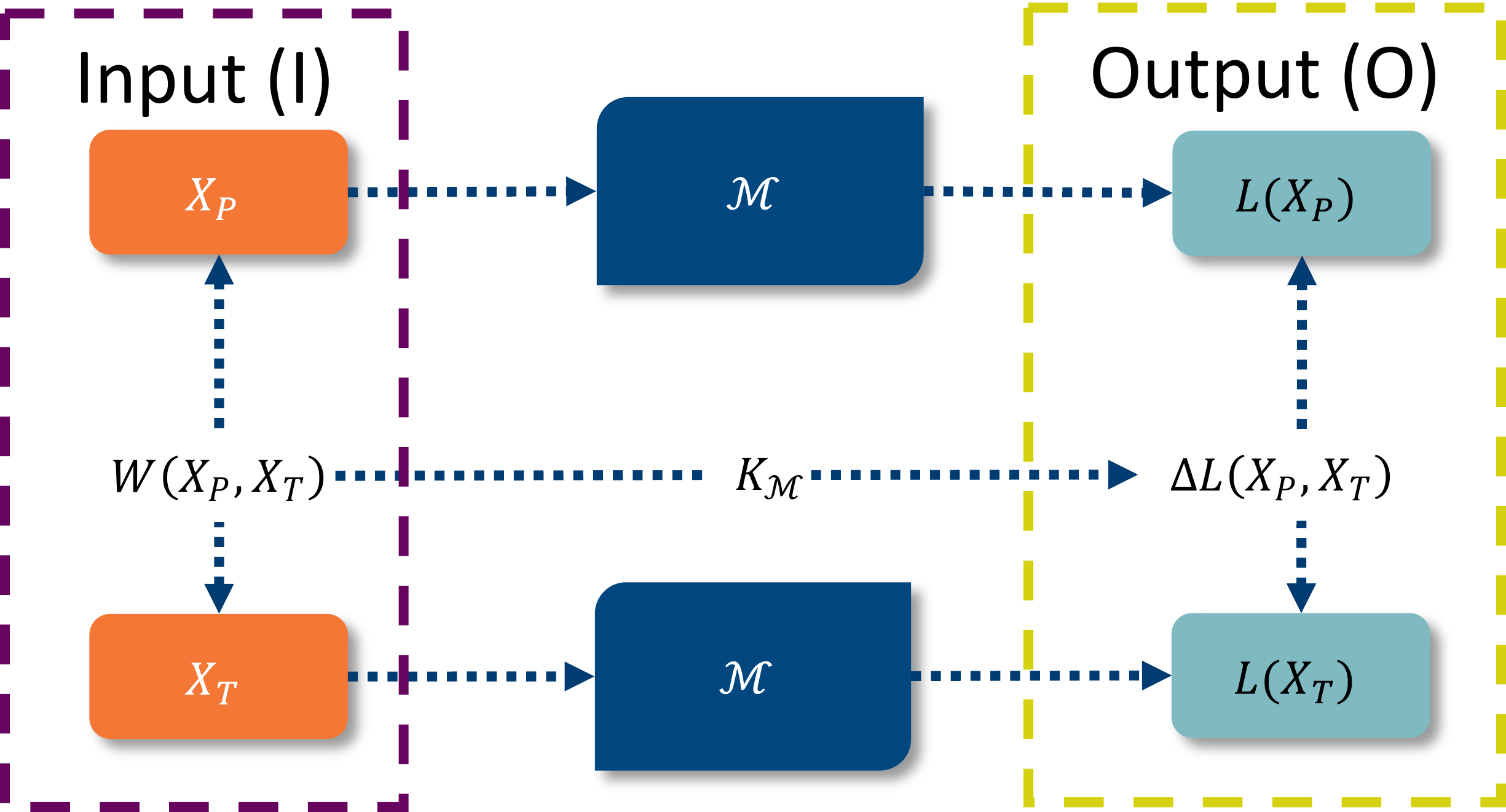}
    \caption{Overview of Proposed Valuation Framework}
    \label{fig:val_io} 
\end{figure}

To provide these performance guarantees we consider the class of Lipschitz continuous performance metrics. Many common loss functions, such as the mean pinball loss (MPL) and MSE are Lipschitz continuous either for any input or a bounded input space \citep{Shwartz2014}. We need to ensure the loss function is Lipschitz continuous in both its data input and its parameters. This can be seen as a form of regularisation, which requires a constrained optimisation procedure \citep{Gouk2021}. As such, requiring Lipschitz continuity is not necessarily restrictive and provides desirable generalisation properties \citep{Lopez2018}. 
\begin{theorem}[Lipschitz Bound]
Given a $K_{\mathcal{M}}$-Lipschitz loss function, $l(x_i)$, for a task $\mathcal{M}$, the difference in the expected loss obtained using $X_P$ or $X_T$ is bounded by the WD between them \citep[adapted from][]{Ghorbani2020}:
\begin{align}
    \lvert L_{\mathcal{M}}(X_P) - L_{\mathcal{M}}(X_T) \rvert \leq K_{\mathcal{M}} \cdot W(X_P,X_T)    
\end{align}
\label{lm:lipschitz}
\end{theorem}
A proof can be found in Appendix \ref{app:proof_lip}. Although the definitional equivalence in (\ref{eq:wass_dual}), upon which Theorem \ref{lm:lipschitz} is based, relates specifically to the WD, the ability to develop such bounds can be extended to other distances using, for example, relationships between distances \citep[see Figure 1 in ][]{Gibbs2002}. Indeed, a connected line of work on developing theoretical performance guarantees for data-driven decision making in non-I.I.D. settings, has proposed such bounds based on KS and TVD \citep{Besbes2022}. Unlike WD based bounds, these are dependent on the diameter of the probability space and may therefore be looser in general. 

\subsection{Effect of Differential Privacy}\label{sec:metric_dp}
Next, we consider the endogenous modelling of the effect of DP on data value. The noise introduced by DP alters the data and therefore affects it's value. We can capture this through upper bounds on the WD \citep{Chhachhi2023}:
\begin{align}
    W(X_i + X_{DP}, X_T) \leq W(X_i, X_T) + W(X_{DP}, \mathfrak{\delta}_0)
    \label{eq:ind_wd}
\end{align}
where, $X_{DP}$ is the distribution of the additive noise mechanism used to achieve DP and $\mathfrak{\delta}_0$ is the Dirac delta distribution concentrated at 0. The first term of the rhs is the WD without noise addition. The second term is $\frac{\Delta_i}{\epsilon_i}$ for the Laplace mechanism and $\frac{2\Delta_i}{\epsilon_i}\sqrt{\frac{\ln(1.25/\delta^{DP}_i)}{\pi}}$ for the Gaussian mechanism. Here, $\Delta_i, \epsilon_i$, and $\delta^{DP}_i$ are the local sensitivity, individual privacy budget and probability of failure, respectively. 

\subsection{Efficient Approximation - Hoeffding Bound}\label{sec:metric_hoeffding}
So far we have shown that the WD is an appropriate measure of data value and that we can provide task-specific performance guarantees. Importantly, we achieve this without having to run the specific task or set of tasks the data may be used for. We only require the computation of the WD of the procured dataset and as such, decouple the valuation process from the complexity of the task. However, computing the WD for any potential subset of data, $P$, remains computationally intensive, as there are $2^{N}-1$ combinations. We therefore introduce an approximation scheme using the Hoeffding Bound and leverage the aggregation effects of our setting.

\begin{theorem}[Hoeffding Bound]
Given a target distribution, $X_T$, an aggregation of $N$ data sources $X_1,\dots,X_N$, the WD between any subset distribution $X_P$ (with $P\subseteq \mathcal{N}$) and the target distribution is bounded, for a given confidence level $\delta$, by: 
\begin{align}
    t_{\delta,N}(P) \leq \sqrt{\frac{\left(\frac{N-|P|}{N}\right) \sum_{i\in P}W_i^2 \ln\left( \frac{2}{1-\delta}\right)}{2|P|^2}}
\end{align}
\label{th:hoeffding}
where $W_i = W(X_i,X_T)$, are the individual WDs for each data source.
\end{theorem}

A full proof is provided in Appendix \ref{app:proof_hoef}. For settings in which the data owners in the market only represent a small proportion of the total population constituting the target distribution ($N << \lvert\mathcal{N}\rvert$) it may be appropriate to adopt an infinite population assumption resulting in a bound without the finite population correction factor $\left(\frac{N-|P|}{N}\right)$ (see (\ref{eq:hoef_inf}) in Appendix \ref{app:proof_hoef}). These probabilistic bounds provide an efficient approximation scheme, decoupled from task/model complexity, and only requires the computation of $N$ individual WDs.

\subsection{Private Computation of the Wasserstein Distance}\label{sec:wd_priv}
Finally, we consider the computation of the WD itself. One of the main motivations of using the WD is to overcome the need to share data owners' raw datasets during the valuation process. As such, we need to ensure that the WD itself is computed in a privacy-preserving manner. Depending on the definition of the datasets this can be achieved using existing PPTs. \cite{Chhachhi2023} showed that when the data under consideration are within the same location-scale family we can obtain closed-form representations of the WD in terms of distributional parameters. As a result, the computation of the WD is equivalent to calculating aggregate sums of parameters. 
This can be efficiently calculated using one or a combination of PPTs such as DP or Multi-Party Computation (MPC). For empirical data, where placing distributional assumptions may be undesirable, the WD between two discrete one dimensional distributions can be calculated privately and efficiently using MPC as a Private Set Intersection - Cardinality problem \citep{Justicia2020}. Importantly, these are again independent of the complexity of the task the data may be used for.

\section{Data Procurement Mechanism}\label{sec:procure}
Having developed the WD-based valuation framework, we now shift our attention to using it to develop a procurement mechanism. We propose three procurement mechanisms for different scenarios. For task-agnostic data procurement we propose a budget feasible mechanism. For task-specific data procurement we propose two mechanisms: an endogenous budget feasible mechanism; and a joint optimisation mechanism, which optimises both data value and payments. We start by formalising the modelling framework which is common to the three proposed mechanisms. Following this we detail the differing objectives and budget constraints of each proposed mechanism.

\subsection{Modelling Framework}
We adopt a Bayesian IMD approach with most of the analysis for a Bayesian optimal mechanism, as detailed in \cite{Ensthaler2014} and \cite{Fallah2022}, being directly applicable to our proposed mechanism. For completeness and notational consistency, we present our adaptation of the modelling framework and relevant results from these papers.

\subsubsection{Data Owners}
Data owners have private data, $X_i$, and a corresponding private reserve price for it, $\theta_i \in \left[\underline\theta_i, \bar\theta_i \right]$, with $0 \leq \underline\theta < \bar\theta < \infty, \quad \forall i$. The reserve price vector is defined as $\theta \coloneqq (\theta_1,\dots,\theta_N)$ on a joint probability space $\Theta \coloneqq \left[\underline\theta_i, \bar\theta_i \right] \times \dots \left[\underline\theta_N, \bar\theta_N \right]$. We assume that the data owner's valuation is drawn from a distribution with a PDF $f_i(\cdot)$ and corresponding CDF $F_i(\cdot)$. In addition, we assume the distributions of $\theta_i$ are independent but not necessarily identically distributed. In this setting, data owner $i$ with reserve price $\theta_i$ receives a payment $t_i$ with probability $q_i$. Their resulting utility is therefore:
\begin{align}
    u_i = t_i - \theta_iq_i
\end{align}

Each owner has a given privacy requirement, $\epsilon_i$, which must be fulfilled if their data is procured. The value of each owner's data is differentiated by their individual WD, $W_i$, which is given by the rhs of (\ref{eq:ind_wd}). We denote the vector of all individual WDs as $W \coloneqq [W_1, \dots, W_N]$.

\subsubsection{Data Buyer}
We assume there is a single data buyer procuring data from the owners, in order to obtain the target distribution $X_T$. In the exogenous budget mechanism, the buyer has a budget $B$. In the endogenous budget and joint optimisation mechanism, the buyer has some reference data $X_R$ (e.g., public dataset) available to them and a corresponding benchmark performance value of their model/task $B_{\mathcal{M}}(X_R)$ using this reference data. We assume the performance metric, $L_{\mathcal{M}}(\cdot)$, is $K_{\mathcal{M}}$-Lipschitz.

In addition, the buyer must set their risk preferences by choosing a confidence level $\delta$ for the Hoeffding Bound in Theorem \ref{th:hoeffding}. Specifically, $\delta$ determines the probability that the Hoeffding bound is greater than the actual WD. For the endogenous budget and joint optimisation mechanisms this translates to the probability of budget feasibility. Since the WDs are calculated privately, as detailed in Section \ref{sec:wd_priv}, and the raw data $\mathcal{X}$ is not shared with the platform prior to procurement, the buyer could be the platform and need not be a TTP.

\subsubsection{Data Acquisition Platform}
The platform receives from each data owner their; $W_i$, reserve prices, $\theta_i$, and privacy budgets, $\epsilon_i$. We assume that there is no known statistical relationship, between $W$ and $\theta$, which the platform could exploit. The platform's task is to determine which owners' data to buy, $q_i$, and how much to pay them, $t_i$, to maximise the benefit. Formally:
\begin{definition}[Data Procurement Mechanism]
    We define a direct mechanism as a tuple $(q,t,V)$ where:
\begin{itemize}
    \item For all $i \in \mathcal{N}, q: \Theta \to (0,1)^N$ is a function which maps reserve prices $\theta$ to a selection probability $q_i$.
    \item For all $i \in \mathcal{N}, t: \Theta \to \mathbb{R}^N_+$ is a function which maps reserve prices $\theta$ to payments $t_i$.
    \item $V(\cdot): W \to \mathbb{R}^+$ is a function which maps vector $W$ to the expected value of a subset of data, $V(\cdot)$.
\end{itemize}   
\end{definition}

\begin{table}[htb]
    \centering
    \caption{Summary of Proposed Data Procurement Mechanisms}
    \begin{tabular}{ccccc}
    \toprule
        & Objective &Budget  & Inputs &  Use Case\\
    \midrule
    \makecell{Exogenous\\Budget}   &  $V(W), \delta$  & $B$ & $\theta, W, B, \delta$ & \makecell{Task-agnostic \\ procurement}\\
    \makecell{Endogenous\\ Budget}& $V(W,K_{\mathcal{M}}, \delta)$ & \makecell{$B_{\mathcal{M}}(X_R) - V(\cdot)$} & \makecell{$\theta, W ,K_{\mathcal{M}},$\\$B_{\mathcal{M}}(X_R), \delta$} & \makecell{Task-specific\\ welfare\\ maximisation}\\
    \makecell{Joint\\ Optimisation} & $V(W,K_{\mathcal{M}}, \delta) + \sum t_i$ & \makecell{$B_{\mathcal{M}}(X_R) - V(\cdot)$} & \makecell{$\theta, W, K_{\mathcal{M}}, $\\$B_{\mathcal{M}}(X_R), \delta$}& \makecell{Task-specific\\ profit\\ maximisation}\\
    \bottomrule
    \end{tabular}
    \label{tab:ass_mech}
\end{table}

\begin{figure}[htb]
    \centering
    \subfloat[Exogenous Budget]{\includegraphics[width=0.4\columnwidth]{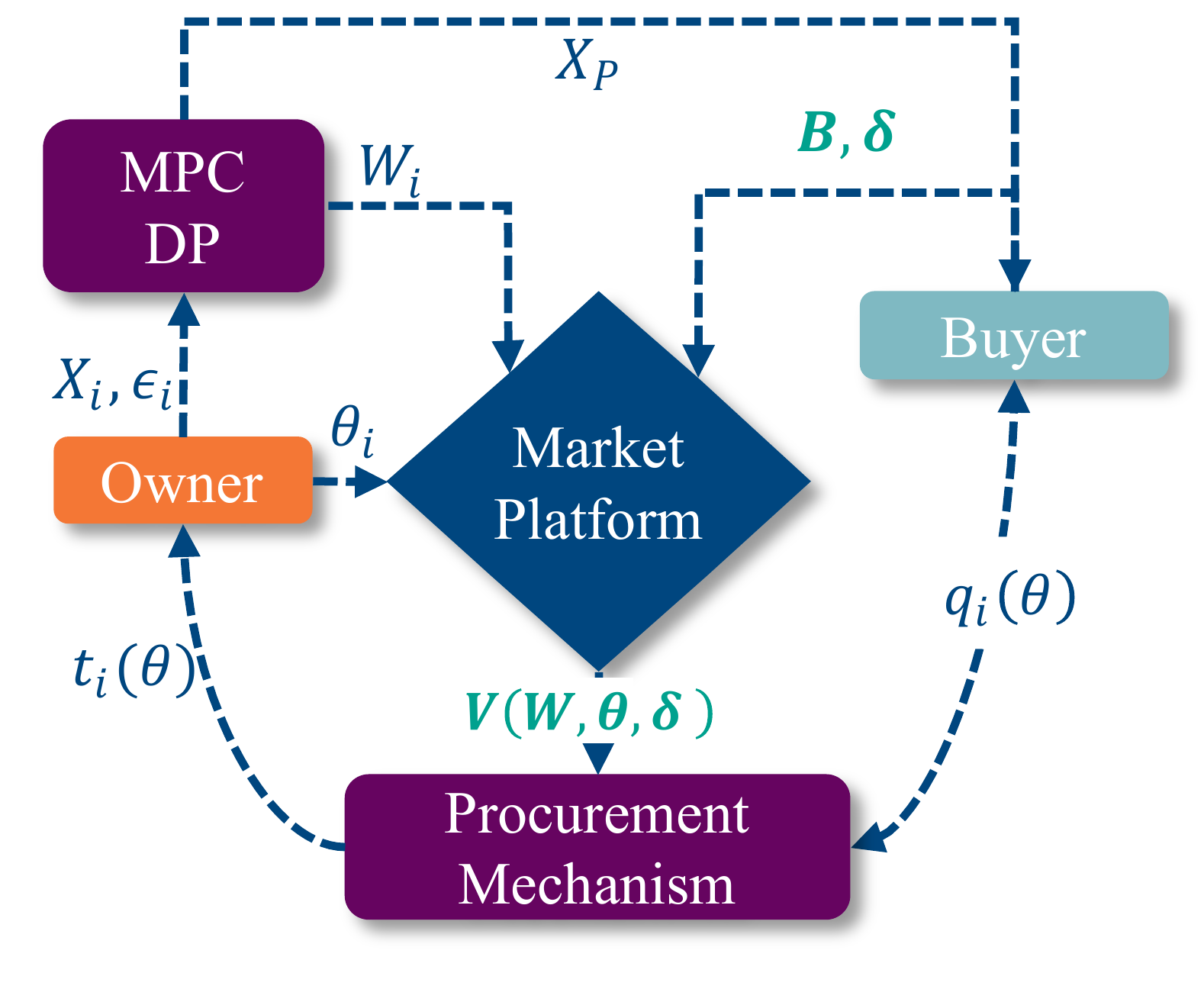}}
    \hfill
    \subfloat[Endogenous Budget/Joint Optimisation]{\includegraphics[width=0.4\columnwidth]{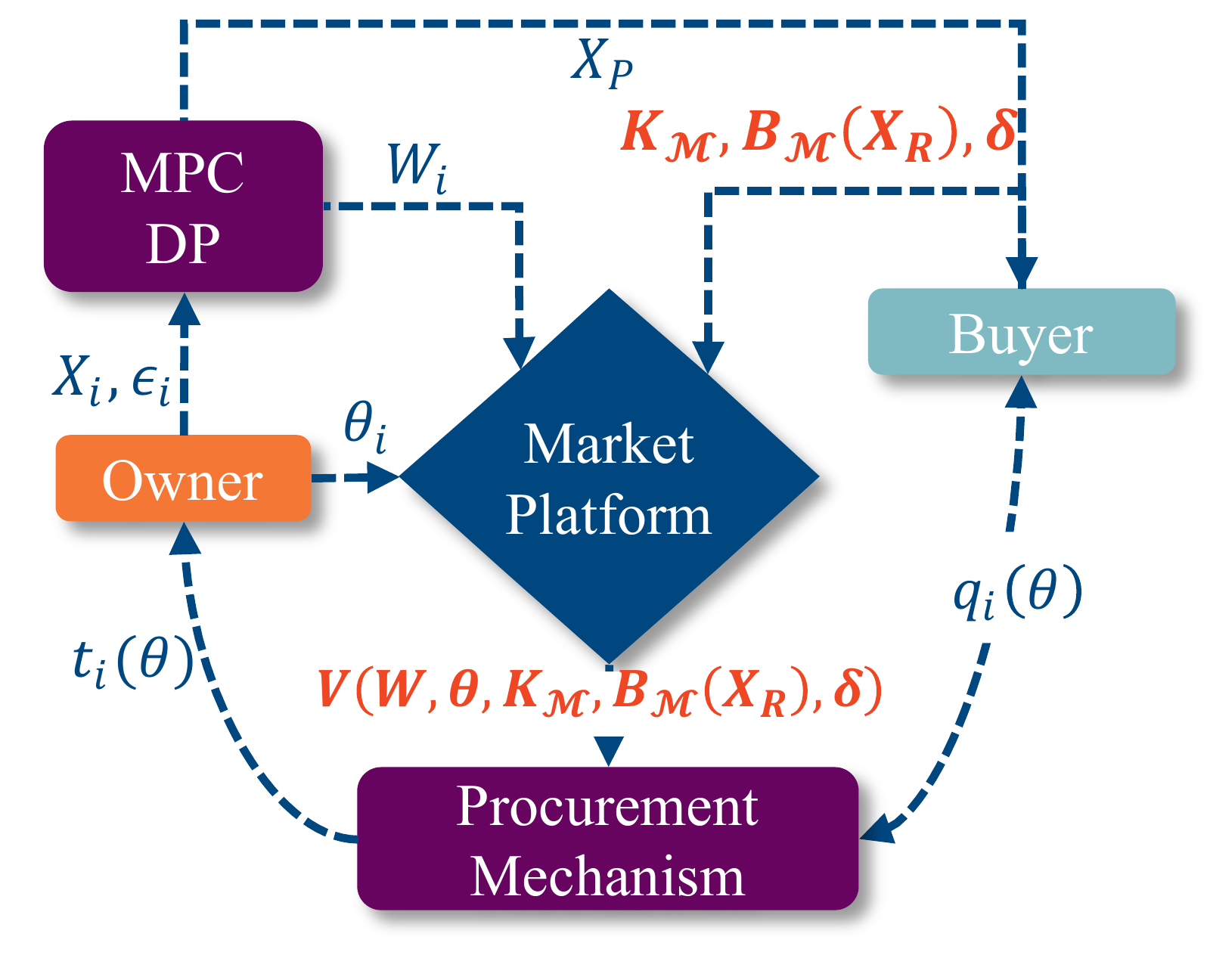}}
    \caption{Dataflow for Proposed Data Procurement Mechanisms}
    \label{fig:market_prop}
\end{figure}

The platform runs a data procurement mechanism to select and remunerate data owners while optimising the buyers' aim. We summarise the objectives, budget constraints, inputs and use cases for each of the proposed mechanisms in Table \ref{tab:ass_mech}. For the exogenous mechanism, the buyer aims to minimise $V(\cdot)$ subject to the external budget, $B$. This models the task-agnostic procurement of data, where $V(\cdot)$ represents how close the procured data is to the target data distribution. This could represent a research institution with a fixed grant aiming to obtain a representative sample which will then be used to a variety of tasks. The objective remains the same in the endogenous budget case, however, the budget is now dependent on $V(\cdot)$. This represents a scenario where the buyer aims to maximise welfare (minimise $V(\cdot)$) while ensuring welfare gains, through data procurement, are commensurate with the associated procurement costs as well as ensuring costs do not exceed a reference budget, $B_{\mathcal{M}}(X_R)$. A potential application for this mechanism would be a data procurement mechanism within an energy collective model, where the buyer would be the community manager, a central coordinating, non-profit entity \citep{Moret2019}. In the joint optimisation case, the buyer aims to minimise $V(\cdot)$ and the associated data costs subject to the same budget as the endogenous case. This models a buyer aiming to maximise their total gains for tasks where data will improve performance, for example, an energy retailer maximising their profits from energy and smart meter data procurement. The information flow for these scenarios is depicted in Figure \ref{fig:market_prop}, with differences in inputs highlighted. 

As we take a Bayesian approach to the mechanism design problem, we also assume that the platform has access to the distribution of owner valuations, $f_i(\theta_i)$. This could be obtained through willingness-to-pay estimates from, for example, stated preference survey studies\citep{Acquisti2016}. We restrict ourselves to deterministic mechanisms, i.e., once reserve prices are reported the mechanism determines, with certainty, which data has been procured. This is motivated by the fact that data owners are interested in their ex-post rather than their expected payments. As such, even for a stochastic mechanism we would need to ensure payments are sufficient for participation for all potential outcomes, otherwise the platform would need to re-adjust payments ex-post \citep{Jarman2017}. As such, $q$ is in fact a vector of binary selection decisions. As the revelation principle applies, we focus on direct revelation mechanism \citep{Jarman2017}. We desire ex-post IC to ensure this equilibrium, by requiring that each owner has no incentive to misrepresent their reserve prices when others report truthfully. In addition, we aim to provide ex-post IR to ensure participation does not leave any data owner worse off, in utility terms. Lastly, the platform should provide ex-interim budget feasibility constraints (BF) to ensure that payments made to data owners do not exceed the budget in expectation. We argue that an ex-interim budget constraint is reasonable in our setting, as the buyer will be procuring data repeatedly. In addition, where budgets are derived based on task-specific performance (the endogenous budget and joint optimisation mechanisms), we envisage the task itself to be stochastic in nature, with data procurement reducing uncertainty.

\subsection{Problem Formulation}
Having defined the parameters of our mechanism, we now develop the platform's task as an optimisation problem. Let $q_i(\theta)$ and $t_i(\theta)$ denote the $i$-th components of $q$ and $t$, respectively and let the subscript $-i$ denote the vector excluding the $i$-th component. Although the platform's problem is similar for the three mechanisms, we start by highlighting the differences in formulation.

\paragraph{Exogenous Budget}
In the exogenous budget mechanism, the platform's problem can be formulated as:
\begin{subequations}
    \begin{align}
        \min_{q,t}  & \int_{\Theta} V (W, q(\theta)) f(\theta)d\theta\tag{\theparentequation}\label{opt:bf}\\
        \text{s.t. }  & U_i(\theta_i|\theta_i) \geq 0, \quad \forall i\in\mathcal{N}, \forall\theta_i \label{eq:ir_cons} \\
        & U_i(\theta_i|\theta_i) \geq U_i(\theta_i'|\theta_i), \quad \forall i \in\mathcal{N}, \forall \theta_i, \theta_i' \label{eq:ic_cons}\\
        & \int_{\Theta} \sum_{i\in \mathcal{N}} t_i(\theta)f(\theta)d\theta \leq B\label{eq:bf_cons}
    \end{align}
\end{subequations}
where, $U_i(\tilde\theta_i|\theta_i) \coloneqq \int_{\Theta_{-i}} t_i(\tilde{\theta_i},\theta_{-i}) f_{-i}(\theta_{-i})d\theta_{-i} -  \int_{\Theta_{-i}}\theta_iq_i(\tilde{\theta_i},\theta_{-i})  f_{-i}(\theta_{-i})d\theta_{-i}$, which denotes the expected utility of an owner with reserve price, $\theta_i$, if they report a reserve price, $\tilde\theta_i$, and all other owners report truthfully. 

$V(W)$ depends on the selection probabilities, $q$, as such, the platform aims to minimise the expected $V$, over the joint owner valuation space, $\Theta$. The first constraint (\ref{eq:ir_cons}) represents IR, we ensure that when data owner $i$ reports their true reserve price, $\theta_i$, their utility must be non-negative. The next constraint, (\ref{eq:ic_cons}), encodes IC. Here we ensure that for an owner with a true reserve price, $\theta_i$, their utility when reporting some other reserve price $\theta_i'$ is no better than that which is achieved when reporting truthfully. Finally, (\ref{eq:bf_cons}) describes the BF constraint.

\paragraph{Endogenous Budget}
The platform's problem for the endogenous budget mechanism is identical to (\ref{opt:bf}) expect (\ref{eq:bf_cons}) is replaced by: 
\begin{align}
    \int_{\Theta} \sum_{i\in \mathcal{N}} t_i(\theta)f(\theta)d\theta \leq B_{\mathcal{M}}(X_R) - \int_{\Theta} V(W,q(\theta), K_{\mathcal{M}},\delta) f(\theta)d\theta
\end{align}
We note that the dependence of $V$ on $K_{\mathcal{M}}$ does not affect the problem structure as it is a scaling factor. The dependence on $\delta$ will be discussed in the following section.

\paragraph{Joint Optimisation}
Finally, for the joint optimisation mechanism, we modify the endogenous budget problem by introducing the expected payments into the platform's objective and budget constraint, resulting in the following objective function:
\begin{align}
    \min_{q,t}  &  \int_{\Theta}  \left[V(W,q(\theta), K_{\mathcal{M}},\delta) + \sum_{i\in \mathcal{N}}t_i(\theta) \right]f(\theta)d\theta
\end{align}
We see that the IC constraint results in an infinite dimensional problem, as we need to ensure it holds for any reserve price realisations within the joint support, $\Theta$.

\subsection{Platform Problem Reformulation}
This section details the reformulations required to obtain a tractable problem for the joint optimisation mechanism. We omit the reformulation for the exogenous and endogenous budget mechanism for the sake of brevity, however, these are obtained using near identical steps.

\subsubsection{Myerson's Lemma}
First, as noted in \cite{Ensthaler2014}, the IC and IR constraints in both problems are identical to those of a buyer in the standard single-item auction problem \citep{Myerson1981}. Assuming the WD of each data source is independent of $\theta_i$, we can apply Myerson's Lemma directly. As a result, we can characterised the payments, $t$, in terms of the selection probabilities, $q$, and the platform's problem is now:
\begin{subequations}
    \begin{align}
        \min_{q} & \ \mathbb{E}_{\Theta}\left[ V(W,q(\theta), K_{\mathcal{M}},\delta) + \sum_{i\in\mathcal{N}}q_i(\theta_i)\psi_i(\theta_i)  \right] \tag{\theparentequation}\\
        \text{s.t. } & Q_i(\theta_i)  \geq Q_i(\widetilde\theta_i),  \quad \forall i\in \mathcal{N},\theta_i, \widetilde{\theta_i}, \theta_i < \widetilde{\theta_i} \label{eq:monotone}\\
        & T_i(\theta_i) = \psi_i (\theta_i) , \quad \forall i \in \mathcal{N} \label{eq:payment_id}\\
            & \mathbb{E}_{\Theta}\left[ V(W,q(\theta), K_{\mathcal{M}},\delta) + \sum_{i\in\mathcal{N}}q_i(\theta_i)\psi_i(\theta_i)  \right] \leq B_{\mathcal{M}}(X_R) \label{eq:bf}
    \end{align}
    \label{opt:myerson}
\end{subequations}
where, $T_i(\theta_i) \coloneqq \int_{\Theta_{-i}} t_i(\theta_i,\theta_{-i})f_{-i}(\theta_{-i})d\theta_{-i}$, $Q_i(\theta_i) \coloneqq \int_{\Theta_{-i}} q_i(\theta_i,\theta_{-i})f_{-i}(\theta_{-i})d\theta_{-i}$, and $\psi_i (\theta_i) = \theta +
\frac{F_i(\theta)}{f_i(\theta)}$, is the virtual cost of data owner $i$. The first constraint, (\ref{eq:monotone}), is the monotonicity requirement for the selection rule, ensuring the selection probability is greater when reporting the true reserve price, $\theta_i$, when reporting a false reserve price, $\bar \theta_i$, if  $\theta_i \leq \bar \theta_i$. Constraint (\ref{eq:payment_id}) is the payment rule, and (\ref{eq:bf}) is the BF constraint in terms of virtual costs.
\subsubsection{Objective Reformulation}
Next, we characterise the function $V(W,q(\theta), K_{\mathcal{M}},\delta)$. The platform aims to select a subset of data $X_P$ which minimises performance loss compared to the target data $X_T$, while ensuring that we do not exceed the budget $B_{\mathcal{M}}(X_R)$. First, we obtain an upper bound on performance using the bounds from Section \ref{sec:wd_motiv}. 
\begin{align}
    V(W,q(\theta), K_{\mathcal{M}},\delta) & = L_{\mathcal{M}}\left(\frac{1}{\lvert P\rvert} \sum_{i\in P}q_i(\theta_i)X_i\right)  - L_{\mathcal{M}}\left(X_T\right) \\
    & \overset{(a)}{\leq}  K_{\mathcal{M}} W\left(X_P,T\right) \\ 
    & \overset{(b)}{\leq}  C\frac{\sqrt{f(\cdot)\sum_{i\in P} q_i(\theta_i)W_i^2}}{\sum_{i\in P}q_i(\theta_i)}
\end{align}
where, (a) results from Theorem \ref{lm:lipschitz}, and (b) results from Theorem \ref{th:hoeffding}. $C$, is a constant dependent on $K_{\mathcal{M}},\delta$ and $N$, and $f(\cdot)$, is a function dependent on $N$ and $q$. They differ depending on whether we assume an finite or infinite population for the Hoeffding bound. 

When the data available to purchase is too expensive and/or of insufficient quality, the buyer will default to their reference data, $X_R$ and their performance will be $B_{\mathcal{M}}(X_R)$. As such, the objective is reformulated as:
\begin{align}
    &\min\left(\mathbb{E}_{\Theta}\left[ V(W,q(\theta), K_{\mathcal{M}},\delta) + \sum\limits_{i\in\mathcal{N}}q_i(\theta_i)\psi_i(\theta_i)\right], B_{\mathcal{M}}(X_R)\right)
\label{opt:min_min}
\end{align}

Finally, to model the minimum in (\ref{opt:min_min}) we introduce an additional selection probability $q_0$, which represents the probability of not buying any data from the data owners and instead relying solely on the reference data, $X_R$. If $q_0 = 0$ then $\sum_{i \in \mathcal{N}}q_i \geq 1$, which indicates the platform has chosen to procure at least one dataset. Conversely, if $q_0 = 1$ then $\sum_{i \in \mathcal{N}}q_i = 0$, which indicates that the platform chooses not to buy any additional data. We assume here that the reference data $X_R$ is available at zero cost, although reference data costs could easily be included with an addition term, $q_0t_0$.

\subsubsection{Point-wise Optimisation}\label{sec:misocp}
We now aim to obtain a point-wise optimisation problem following the approach of \cite{Fallah2022}. The required reformulations depend on our population assumptions for the Hoeffding Bound, which determine $C$ and $f(\cdot)$. For the sake of brevity, we only present the reformulation assuming a finite population here\footnote{The MISOCP formulation under the infinite population Hoeffding bound can be found in Appendix \ref{app:inf_misocp}.}. In this case $C^{FIN} = K_{\mathcal{M}}\sqrt{\frac{\ln\left(\frac{2}{1-\delta}\right)}{2(N-1)}}$ and $f(\cdot) = N - \sum_{i\in\mathcal{N}}q_i$. Ignoring the monotonicity requirement in (\ref{eq:monotone}), the platform problem becomes:
\begin{subequations}
    \begin{align}
    \begin{split}
        \min_{q} \quad& C^{FIN}\sqrt{\frac{\left(N-\sum_{i \in \mathcal{N}} q_i\right)\sum^{N}_{i=1} q_iW_i^2}{\left(\sum_{i \in \mathcal{N}} q_i\right)^2}} 
        + \sum_{i \in \mathcal{N}} q_i \psi_i(\theta_i) + q_0 B_{\mathcal{M}}(X_R)
    \end{split} \tag{\theparentequation}\label{obj:fin}\\
        \text{s.t.}\quad &  1\leq \sum^{N}_{i=0} q_i \leq N\label{obj:fin1}\\
        & t_i = q_i \psi_i, \quad \forall i \in \mathcal{N}\label{obj:fin2}\\
        & q \in \{0,1\}^{N+1}\label{obj:fin3}
    \end{align}
    \label{opt:fin_or}
\end{subequations}
In order to convexify (\ref{obj:fin}), we introduce a number of auxiliary variables and substitutions. First, note that $N - \sum_{i=1}^{N} q_i = \sum_{i=1}^{N}(1-q_i)$, resulting in the numerator within the square root being $\sum_{i=1}^{N}W_i^2q_i\left(\sum_{j=1}^{N}(1-q_j)\right)$. The binary products $q_i\sum_{j=1}^{N}(1-q_j)$ are linearised by introducing auxiliary binary variables $r_{i,j}$ and constraints (\ref{eq:fin_binlin1})-(\ref{eq:fin_binlin3}). The resulting objective term is $\sqrt{\sum_{i=1}^{N} \sum_{j\neq i} W_i^2 r_{i,j}}$. Note that as $q_i(1-q_i) = 0$, we only require $N^2-N$ auxiliary binary variables. Lastly, $\sqrt{\sum_{i=1}^{N} \sum_{j\neq i} W_i^2 r_{i,j}}$ is equivalent to a matrix norm (as $r$ is binary) which, can be reformulated as a SOC constraint in (\ref{eq:fin_cone}). This is achieved by introducing $s$ to linearise the objective, and $z$ and constraints  (\ref{eq:fin_bincon1})-(\ref{eq:fin_bincon2}) to linearise the resulting binary-continuous products. The resulting MISOCP is:
\begin{subequations}
    \begin{align}
        \min_{\substack{q,r,s,z}} \quad&   q_0 B_{\mathcal{M}}(X_R) + s  + \sum_{i \in \mathcal{N}} q_i \psi_i(\theta_i)\label{eq:misocp_obj}\tag{\theparentequation}\\
        \text{s.t.}\quad & C^{FIN}\lVert W r \rVert \leq \sum_{i \in \mathcal{N}} z_i \label{eq:fin_cone}\\
        & r_{i,j} \leq q_i, \quad \forall i \in \mathcal{N}\label{eq:fin_binlin1}\\
        & r_{i,j} \leq 1 - q_j, \quad \forall j \in \mathcal{N}\\
        & r_{i,j} \geq q_i - q_j, \quad \forall i \in \mathcal{N}, j \in \mathcal{N}/i \label{eq:fin_binlin3}\\
        & 0 \leq z_i \leq  Mq_i, \quad \forall i \in \mathcal{N}\label{eq:fin_bincon1}\\
        & 0 \leq s - z_i \leq  M(1-q_i), \quad \forall i \in \mathcal{N}\label{eq:fin_bincon2}\\
        & \text{Constraints (\ref{obj:fin1}) - (\ref{obj:fin3})}\\
        & r \in \{0,1\}^{N^2-N}, s \in \mathbb{R}_+, z \in \mathbb{R}^{N}_{+}
    \end{align}
    \label{opt:fin_misocp}
\end{subequations}
where, $M \geq \min \left(B_{\mathcal{M}}(X_R),  K_{\mathcal{M}}\sqrt{\frac{\ln\left(\frac{2}{1-\delta}\right)}{2}}\max_i W_i\right)$.

Finally, to ensure feasibility has been maintained we show that the allocation $q$ is monotonically decreasing in $\theta$. The proof can be found in Appendix \ref{app:mono}. We note that the finite population assumption results in an additional $N^2 - N$ binary variables compared to the infinite population assumption (see Appendix \ref{app:inf_misocp}). The valuation and procurement performance implications of this will be discussed through the case studies presented in Section \ref{sec:case}. 

\subsection{Reference Budget}
For all proposed mechanisms the buyer is required to provide an external budget, $B$ or $B_{\mathcal{M}}(X_R)$ and the mechanism ensures budget feasibility with respect to this external budget. In the joint optimisation and endogenous budget mechanisms, we aim to ensure that the expected performance loss, in monetary terms, due to using $X_P$ instead of $X_T$ and the associated payments to procure $X_P$ is less than the performance loss achieved with the (free) reference data, $X_R$. As such, we can define the external budget as $B =  L_{\mathcal{M}}\left(X_R\right) - L_{\mathcal{M}}\left(X_P\right)$. However, as we do not have access to $L_{\mathcal{M}}\left(X_P\right)$, we develop a lower bound on the budget:
\begin{align}
    B & = L_{\mathcal{M}}\left(X_R\right) - L_{\mathcal{M}}\left(X_P\right)  \\
    & = \left[L_{\mathcal{M}}\left(X_R\right) - L_{\mathcal{M}}\left(X_T\right)\right] - \left[L_{\mathcal{M}}\left(X_P\right) - L_{\mathcal{M}}\left(X_T\right)\right] \\
    & \overset{(a)}{\geq} \left[L_{\mathcal{M}}\left(X_R\right) - L_{\mathcal{M}}\left(X_T\right)\right] - K_{\mathcal{M}} W\left(X_P,T\right) \\ 
    & \overset{(b)}{\geq} \left[L_{\mathcal{M}}\left(X_R\right) - L_{\mathcal{M}}\left(X_T\right)\right] -  C(K_{\mathcal{M}},\delta,N)\frac{\sqrt{f(N,q)\sum_{i\in P} q_i(\theta_i)W_i^2}}{\sum_{i\in P}q_i(\theta_i)}
\end{align}
where, (a) results from Theorem \ref{lm:lipschitz}, and (b) results from Theorem \ref{th:hoeffding}.

Ideally, $B_{\mathcal{M}}(X_R) = L_{\mathcal{M}}(X_R) - L_{\mathcal{M}}(X_T)$, however, we do not have access to the $L_{\mathcal{M}}(X_T)$, as this would also violate the data privacy of the owners. The buyer is therefore forced to estimate $B_{\mathcal{M}}(X_R)$, using for example, historical performance, or theoretical problem-specific bounds. We explore the implications of under or over-estimation below:
\begin{itemize}
    \item \textbf{Lower Bound}, if $B_{\mathcal{M}}(X_R) <  L_{\mathcal{M}}(X_R) - L_{\mathcal{M}}(X_T)$, the mechanism retains budget feasibility. As the lower bound results in an under-estimation of the budget, the buyer ends up with less data than they could have procured.
    \item \textbf{Upper Bound}, if $B_{\mathcal{M}}(X_R) >  L_{\mathcal{M}}(X_R) - L_{\mathcal{M}}(X_T)$, the mechanism can no longer provide budget feasibility guarantees. The over-estimation will lead to the buyer purchasing more data than they should. If the resulting performance and payments are higher than $L_{\mathcal{M}}(X_R) - L_{\mathcal{M}}(X_T)$, the buyer will be worse off than if they simply used the reference data. However, as the WD provides an upper bound on the performance loss over-estimation does not necessarily lead to budget infeasibility. A natural choice for an upper bound would be the Lipschitz bound, $K_{\mathcal{M}}W(X_R,X_T)$.
\end{itemize}

In both cases, the cost of estimation error is borne by the buyer, thus incentivising the buyer to produce accurate estimates of $B_{\mathcal{M}}(X_R)$. Data owners, on the other hand, are ensured a payment above their reserve prices, thereby maintaining IR. This ensures a data owner/user-centric approach. If we wish to maintain budget feasibility, we could develop a privacy-preserving protocol to calculate $L_{\mathcal{M}}(X_R) - L_{\mathcal{M}}(X_T)$. The accuracy would be dependent on the technique and the particular performance metric. We note, of course, that such a technique could then be used to create a privacy-preserving CG framework. However, we argue that our approach still provides benefits, in terms of computational costs, in this scenario. A CG still requires the calculation of each coalition value whereas we would only require the calculation of one additional term $L_{\mathcal{M}}(X_T)$. The computational advantages are particularly pronounced when the underlying model is computationally intensive. 

\section{Case Study: Parameter Estimation with Synthetic Data}\label{sec:case}
In order to illustrate the efficacy of our proposed mechanisms, we consider the problem of estimating parameters of synthetic data. We first evaluate the efficacy of the WD based valuation framework against a range of alternatives. We then investigate the performance of the three procurement mechanisms, including benchmarking where applicable. All computations were implemented in Python using CVXPY with Gurobi 9.5.0, on a DELL XPS 15 with an 11th Gen Intel\textregistered Core\texttrademark i7-11800H processor and 64GB RAM\footnote{Our code is publicly available at:  \href{https://github.com/saurabac/Wasserstein-Data-Markets}{https://github.com/saurabac/Wasserstein-Data-Markets}.}.

\subsection{Data Valuation}\label{sec:cs_data_val}
As the aim of using the WD is to provide a task-agnostic valuation metric, we investigate three use cases and associated loss functions, commonly observed within the machine learning literature; (1) mean estimation/RMSE, (2) quantile estimation/MPL (including median/MAE) and (3) newsvendor cost/NV. We test three different data distributions with different properties; Gaussian (symmetric and unbounded), uniform (symmetric and bounded) and exponential (asymmetric and unbounded). We generate $N = 8$ data sources over 50 trials with location, $\alpha \sim U(10,16)$, and scale, $\beta \sim U(1,3)$ parameters. The target distribution is the Euclidean barycenter of the 8 data sources and we calculate the distances and loss function values for all 255 combinations of data sources. 

\subsubsection{Lipschitz Bounds}
\begin{figure}[htb]
    \centering
    \subfloat[Gaussian]{
    \includegraphics[width=0.3\textwidth]{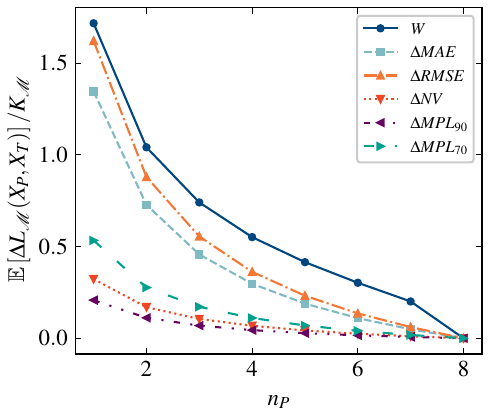}}
    \hfill
    \subfloat[Uniform]{
    \includegraphics[width=0.3\textwidth]{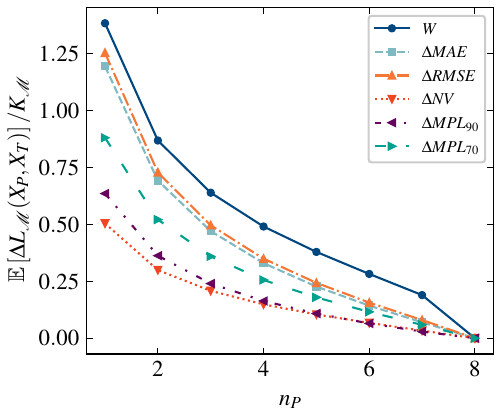}}
    \hfill
    \subfloat[Exponential]{
    \includegraphics[width=0.3\textwidth]{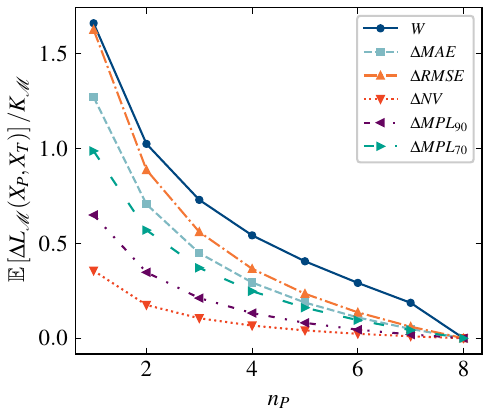}}
    \vfill
    \subfloat[Gaussian]{
    \includegraphics[width=0.3\textwidth]{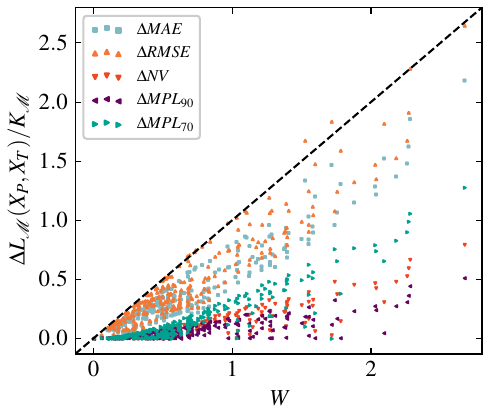}}
    \hfill
    \subfloat[Uniform]{
    \includegraphics[width=0.3\textwidth]{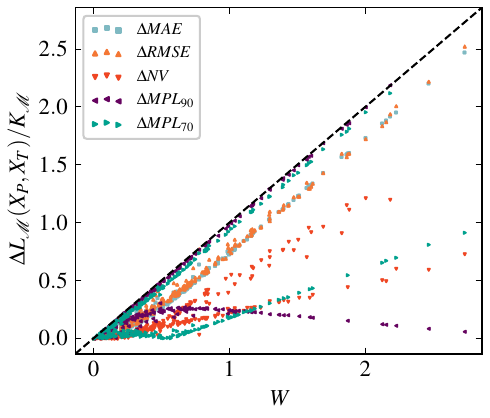}}
    \hfill
    \subfloat[Exponential]{
    \includegraphics[width=0.3\textwidth]{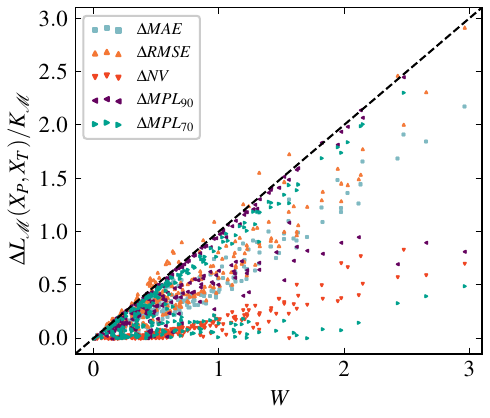}}
    \caption[Performance of Lipschitz Bounds]{Performance of Lipschitz Bounds for Different Data Distributions. Top: Mean of Metrics as Function of Coalition Size. Bottom: Scatterplot of all Coalitions against WD.}
    \label{fig:lip_perf}
\end{figure}

To assess the performance of the Lipschitz bounds, we are interested in the difference $\Delta L_{\mathcal{M}}(X_P,X_T) - K \cdot W(X_P,X_T)$, where, $\Delta L_{\mathcal{M}}(X_P,X_T)$ is the difference between the loss function for task, $\mathcal{M}$, when using a subset of the data, $X_P$, and using the target distribution, $X_T$. The top plots in Figure \ref{fig:lip_perf} show the expected value of the WD, $\mathbb{E}[W]$ (over the 50 trials), and $\Delta L_{\mathcal{M}}(X_P,X_T)$, divided by the associated Lipschitz constant, $K_{\mathcal{M}}$. The bottom plots show the loss function values, again divided by $K_{\mathcal{M}}$, against the WD for a particular trial. The dashed line represents the Lipschitz boundary with values below the line satisfying the bound and values above it violating the bound. From the top plots we see that the Lipschitz bound is tight (the difference is small), on average, for the MAE and RMSE across all distributions. However, for unbounded distributions (Gaussian and Exponential), the RMSE is in fact not Lipschitz, resulting in some coalition values violating the Lipschitz bound, as shown in bottom plots. For the MPL and NV the Lipschitz bound holds in all scenario but is much looser. 

\subsubsection{Task Correlations}
\begin{figure}[htb]
    \centering
    \subfloat[$\rho$ (Gaussian)]{
    \includegraphics[width=0.3\textwidth]{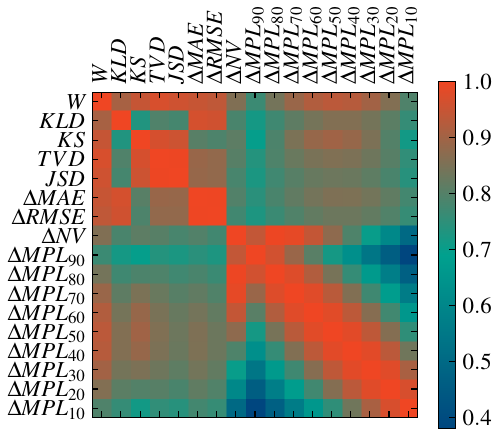}}
    \hfill
    \subfloat[$\rho$ (Uniform)]{
    \includegraphics[width=0.3\textwidth]{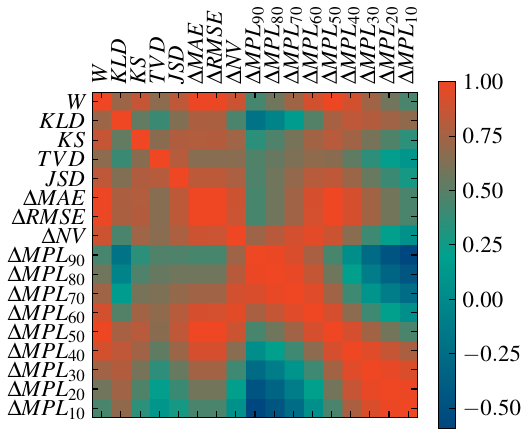}}
    \hfill
    \subfloat[$\rho$ (Exponential)]{
    \includegraphics[width=0.3\textwidth]{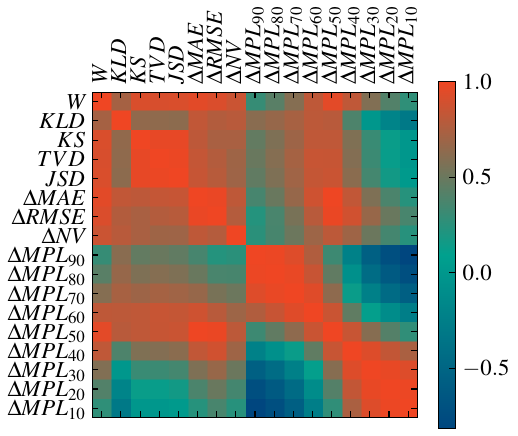}}
    \vfill
    \subfloat[$\rho^W \geq \rho^*$ (Gaussian)]{
    \includegraphics[width=0.3\textwidth]{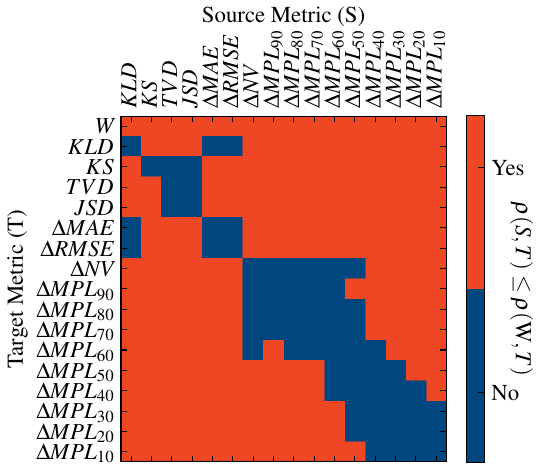}}
    \hfill
    \subfloat[$\rho^W \geq \rho^*$ (Uniform)]{
    \includegraphics[width=0.3\textwidth]{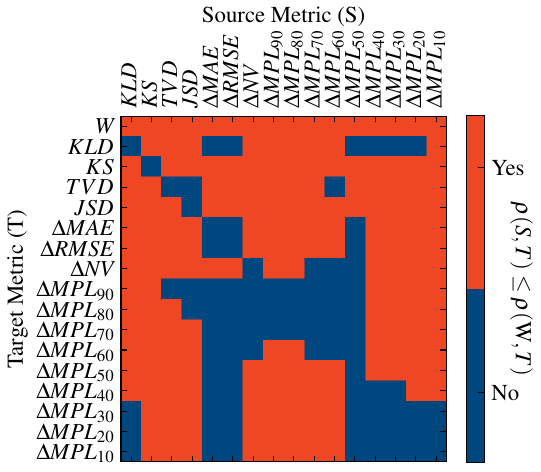}}
    \hfill
    \subfloat[$\rho^W \geq \rho^*$ (Exponential)]{
    \includegraphics[width=0.3\textwidth]{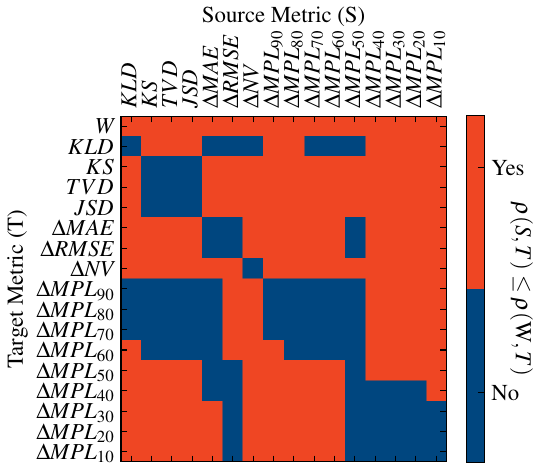}}
    \caption[Correlations between Distances and Loss Functions]{Correlations between Distances and Loss Functions.}
    \label{fig:corr_perf}
\end{figure}

Figure \ref{fig:corr_perf} shows the correlation performance between the distances and loss functions considered. The top plots show the average correlation coefficients $\rho$ over the 50 trials and the bottom plots show whether using the WD results in a higher correlation with a target metric (y-axis) or the another source metric (x-axis) has a higher correlation. We see that the correlation between the WD and the loss functions is between 0.7 and 1.0 for Gaussian data. Overall, we see that the WD has higher correlations for almost all the considered loss functions. The KLD has higher correlations with MAE and RMSE for Gaussian data and the MPL for lower quantiles ($\leq 30$) for Uniform data. For exponential data we see that, for higher quantiles ($\geq 70$), the other distances out-perform the WD. However, we also observe that for certain trials using uniform or exponential data, the KLD is undefined due to non-overlapping supports and/or PDF estimation. This is more pronounced for smaller coalition sizes as these are likely to be further from the aggregate/target distribution. Although we see that the best distance varies, the WD is consistent with high correlations across loss functions and distribution.

\subsubsection{Shapley Allocations}
\begin{figure}[htb]
    \subfloat[Distances]{
    \includegraphics[width=0.3\textwidth]{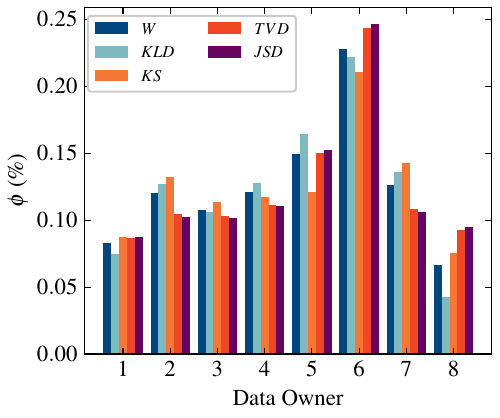}}
    \hfill
    \subfloat[Loss Functions]{
    \includegraphics[width=0.3\textwidth]{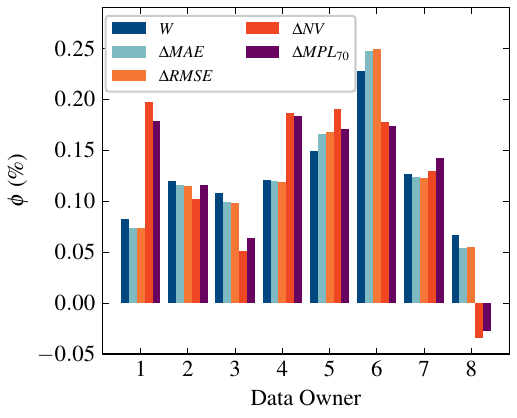}}
    \hfill
    \subfloat[Average Mis-Allocation \label{fig:shapley_diff}]{
    \includegraphics[width=0.3\textwidth]{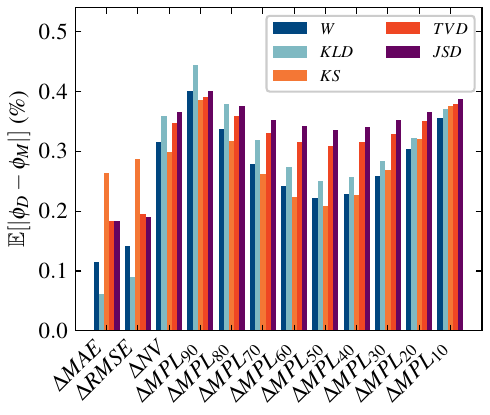}}
    \caption{Shapley Allocations for Gaussian Data.}
    \label{fig:shapley}
\end{figure}

Figure \ref{fig:shapley} shows the Shapley allocation proportions for each data source using different (a) distances and (b) loss functions for Gaussian data for a particular trial. Similar dynamics are observed for uniform and exponential data and are therefore omitted. For the loss functions the characteristic value function used was $V = \max_{s \subseteq \mathcal{N}}(L(X_s))-L(X_P)$. Similarly, for distances the characteristic value function is $V = \max_{s \subseteq \mathcal{N}}(d(X_s,X_T))-d(X_P,X_T)$. We see that across the distances the allocation proportions are quite similar with differences less than 5\%. In contrast, the allocations exhibit significantly more variation across loss functions. Figure \ref{fig:shapley_diff} shows the average differences or mis-allocations using different distances across the loss functions. The KLD performs better for MAE and RMSE and KS is better for higher quantiles. Overall, the WD results in either the least or second least in mis-allocations, suggesting a more stable notion of value. The results are broadly in line with the correlation analysis, that is, higher correlations lead to lower average mis-allocations.

\subsubsection{Hoeffding Bounds}
\begin{figure}[htb]
    \centering
    \subfloat[Expected Value ($\delta$ = 0.95)\label{fig:hoef_perf_avg}]{
    \includegraphics[width=0.3\textwidth]{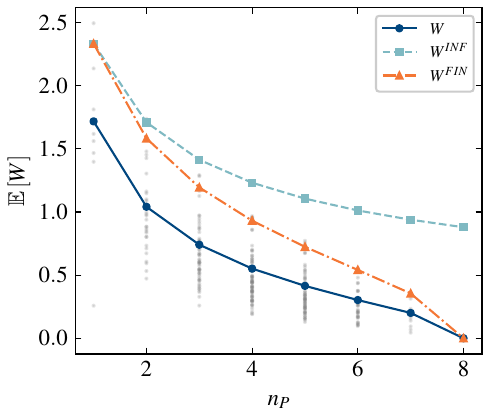}}
    \hfill
    \subfloat[Effect of $\delta$ on $W^{INF}$\label{fig:hoef_perf_inf}]{
    \includegraphics[width=0.3\textwidth]{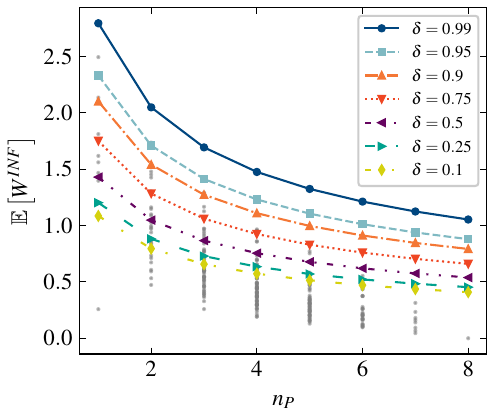}}
    \hfill
    \subfloat[Effect of $\delta$ on $W^{FIN}$\label{fig:hoef_perf_fin}]{
    \includegraphics[width=0.3\textwidth]{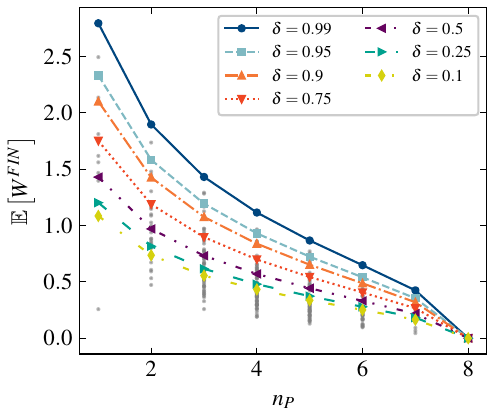}}
    \caption[Hoeffding Bounds for Gaussian Data]{Hoeffding Bounds for Gaussian Data. Grey Dots are Actual WDs for each Potential Coalition.}
    \label{fig:hoef_perf}
\end{figure}

Figure \ref{fig:hoef_perf_avg} shows the expected value of the Hoeffding bounds and the actual WD. The grey dots are the WDs for each coalition of the given size. We include the Hoeffding bounds with ($W^{FIN}$) and without ($W^{INF}$) the finite population correction as these have implications on the complexity of the market formulations, specifically the number of binary variables in the resulting MISOCP. The finite population formulation ensures convergence to zero with a full dataset whereas the infinite formulation maintains a non-zero bias which is more significant for smaller datasets. Figure \ref{fig:hoef_perf_inf} and \ref{fig:hoef_perf_fin} show the effect of tuning the confidence level, $\delta$, on the Hoeffding bounds. 

\begin{figure}[htb]
    \subfloat[Hoeffding Minimiser\label{fig:hoef_opt_mini}]{
    \includegraphics[width=0.3\textwidth]{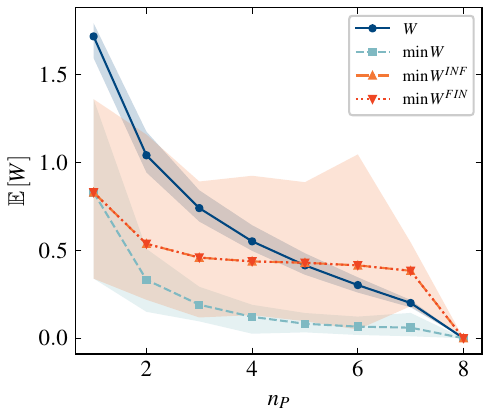}}
    \hfill
    \subfloat[Correlation]{
    \includegraphics[width=0.3\textwidth]{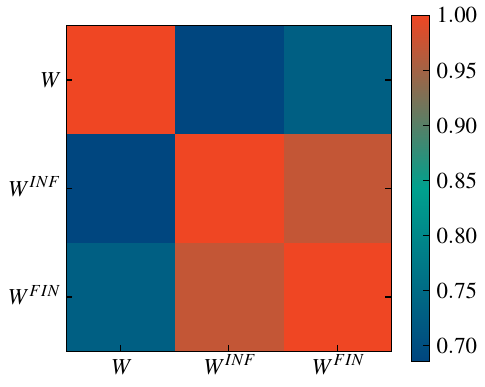}}
    \hfill
    \subfloat[Shapley Allocation]{
    \includegraphics[width=0.3\textwidth]{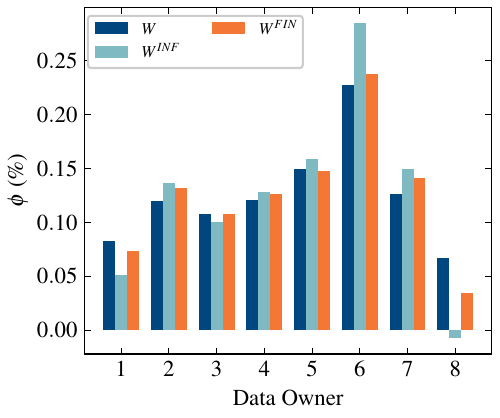}}
    \caption{Hoeffding Bound Performance with $\delta = 0.95$}
    \label{fig:hoef_opt}
\end{figure}

The Hoeffding bound offers an attractive option over which to optimise, capturing aggregation effects without needing to calculate the WD for each combination of data sources. To this end, we compare the average minimisers of the actual WD, achieved when calculating each combination, and the minimisers of the Hoeffding bounds. As shown in Figure \ref{fig:hoef_opt_mini}, using the Hoeffding bounds improves average performance when compared to random selection (the average WD, $W$, for a given coalition size shown in dark blue), when the coalition size is small compared to the total number of data source ($n_P \leq 5$ in this case). As such, access to combinatorial information results in better performance overall, represented by $\min W$. 

The correlations between the actual WD and the Hoeffding approximations are $\rho(W,W^{FIN}) = 0.72$ and $\rho(W,W^{INF}) = 0.67$. Again, we see this also effects the Shapley allocations, although the finite Hoeffding bound provides similar allocations to the actual WD. We also note that the Hoeffding bound approximation results in a bias. Although the Hoeffding bound accounts for the aggregation effects that is the convergence of the data distribution to the target distribution as $n_P \to N$, it does not capture the combinatorial effects of the aggregation itself. For example, given two distributions which are individually far away from the target but when aggregated may be much closer to the target distribution. This establishes the trade-off between computational costs and accuracy for data valuation in our setting.

\subsection{Data Procurement}\label{sec:data_proc}
The three proposed mechanisms serve different purposes and are therefore not directly comparable. As such, we set up two different case studies, the first to assess our exogenous budget mechanism against existing exogenous budget mechanism and the second to assess the endogenous budget mechanisms (including the joint optimisation mechanism) against a centralised benchmark. We use the same synthetic Gaussian data as in Section \ref{sec:cs_data_val}.

\subsubsection{Exogenous Budget Mechanisms}
To evaluate the performance of the proposed exogenous budget mechanism, using the finite ($FIN$) and infinite ($INF$) formulations, we compare them against the following existing approaches and benchmarks:
\begin{itemize}
    \item Central ($CEN$) -  Assuming full access to the value of each coalition the mechanism selects the coalition with maximum value (minimum statistical distance, $d$) that is budget feasible. Provides a benchmark of best possible performance.
    \item Random ($RAND$) - Assume a budget feasible coalition is selected at random resulting in the average value of budget feasible coalitions. Provides a worst case benchmark in the absence of an optimal selection criteria.
    \item Single Minded Query ($SMQ$) \citep{Zhang2021}\footnote{Bayesian mechanism, similar to our exogenous budget mechanism, which aims to maximise (reserve price independent) value, $V$, subject to ex-interim budget feasibility. The offers are determined by solving an auxiliary, convex, optimisation problem, which, for uniformly distributed reserve prices, is a SOCP. Solved using MOSEK due to numerical issues in Gurobi.} - Assumes the value of each data owner is $V_i = \frac{1}{d_i}$ and data owners' values are additive ($V = \sum_i V_i$). This results in a separable problem where each data owner receives a take-it-or-leave-it offer. If the owners' reserve price is lower then the offer, $\theta^*$, the owners' data is purchased.
    \item Greedy Knapsack ($PTAS$) \citep{Ren2022}\footnote{Data owners are sorted, in ascending order, by their cost per unit value ($g_i = \theta_i d_i$). The mechanism then finds the largest index $k$ which satisfies $g_k \leq  \frac{B}{\sum_{i\leq k} \frac{1}{d_i}}$. All owners $i \leq k$, are selected and paid $p_i = \min\left\{\frac{B}{\sum_{i\leq k} \frac{1}{d_i}}, g_{k+1}  \right\}\frac{1}{d_i}$.} – Provides a polynomial time approximation scheme to the same problem as above, in a prior-free environment. 
\end{itemize}

We investigate the performance of the above mechanisms for the different statistical distances discussed in Section \ref{sec:wd_motiv} (WD, KLD, KS, TVD and JSD), different budget levels, and correlations, $\rho(\theta,d) \in \{-1,0,1\}$, between reserve prices and the value metric (distance). Extant literature suggests that consumer valuations of personal data are not necessarily linked to its' actual value but other considerations, such as privacy. We therefore consider the full range of potential correlations. The reserve prices are assumed to follow a uniform distribution ($\theta \sim U(0,\bar\theta)$) and the budget levels are multiples of the maximum reserve price, $\bar\theta = 1$, $B \in \{0.1\bar\theta N, 0.2\bar\theta N,  \dots, \bar\theta N\}$. 

\begin{figure}[htb]
    \centering
    \subfloat[$\rho(W,\theta) = -1$\label{fig:fix_budget_wass_n1}]{
    \includegraphics[width=0.3\textwidth]{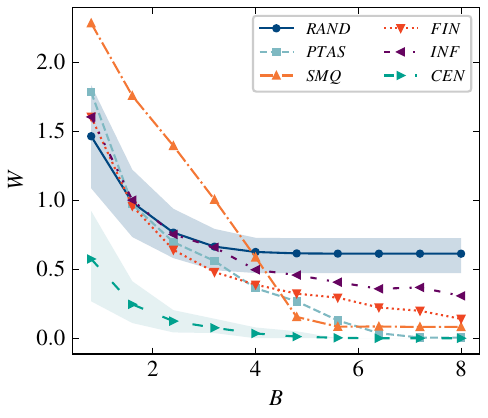}}
    \hfill
    \subfloat[$\rho(W,\theta) = 0$\label{fig:fix_budget_wass_0}]{
    \includegraphics[width=0.3\textwidth]{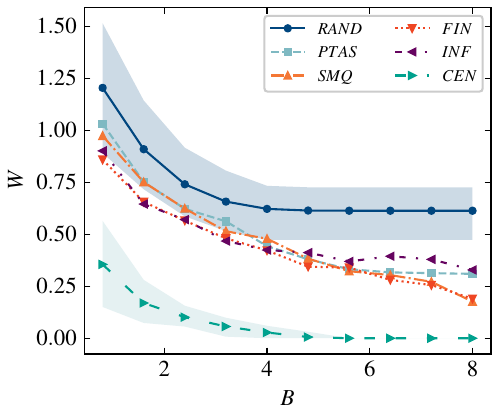}}
    \hfill
    \subfloat[$\rho(W,\theta) = 1$\label{fig:fix_budget_wass_1}]{
    \includegraphics[width=0.3\textwidth]{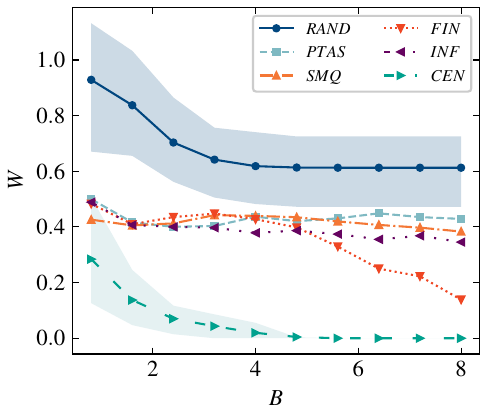}}\\
    \vfill
    \subfloat[$\rho(W,\theta) = -1$\label{fig:fix_budget_wass_num_n1}]{
    \includegraphics[width=0.3\textwidth]{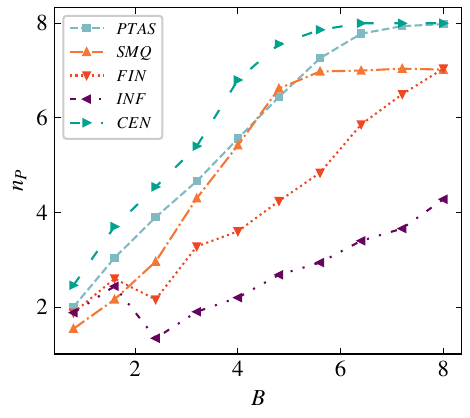}}
    \hfill
    \subfloat[$\rho(W,\theta) = 0$\label{fig:fix_budget_wass_num_0}]{
    \includegraphics[width=0.3\textwidth]{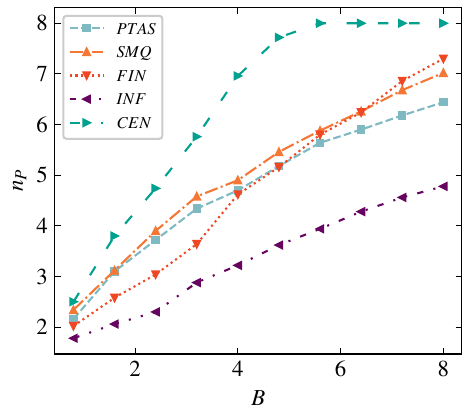}}
    \hfill
    \subfloat[$\rho(W,\theta) = 1$\label{fig:fix_budget_wass_num_1}]{
    \includegraphics[width=0.3\textwidth]{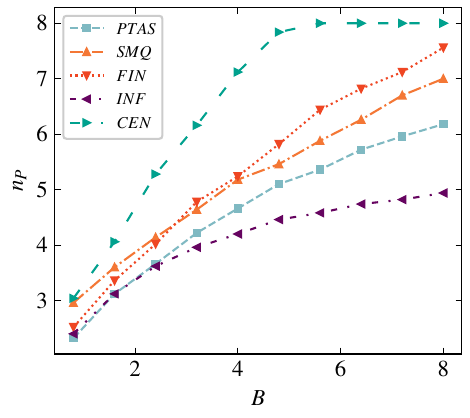}}
    \caption[Exogenous Budget Mechanisms under Different Value-Price Correlation]{Exogenous Budget Mechanisms under Different Value-Price Correlation. Performance (top) and Average Number of Data Owners Selected (bottom).}
    \label{fig:fix_budget_wass}
\end{figure}

\paragraph{Market Mechanisms}
Figure \ref{fig:fix_budget_wass} shows the performance of the different budget feasible mechanisms considered for minimising the WD. The average WD of the selected coalition across the 50 trials is represented by the lines. In addition for the two benchmarks ($CEN$, $RAND$), we include the 95\% confidence intervals. We see that overall, performance improves when the $\rho(W,\theta) = 0$ or $1$. This is expected, as the later scenario assumes owners with a higher WD/lower value have higher costs, resulting in higher value per unit cost. In addition, performance of all mechanisms is generally between the two benchmarks, with the exception of $SMQ$ in the case where WD and cost are negatively correlated. In this case $SMQ$ picks coalitions of smaller size, because of the assumptions used to develop the mechanism. 

$SMQ$ assumes value is additive, resulting in a separable problem, where the aim is to determine individual payment thresholds. The payment thresholds are determined by maximising the expected value, based on the reserve price distributions $f_i(\cdot)$ without considering the actual reserve prices, $\theta_i$. As a result, the mechanism allocates some of the budget to owners which are not selected. The drawback of this effect is most pronounced in a budget constrained scenarios, that is for low $B$ (with $\rho(W,\theta) = -1$ being the extreme case). However, the separability of the problem also allows the mechanism to drop incentive compatibility. As such, it is able to purchase more data, when the budget constraints are higher (less budget is 'wasted' on un-selected owners), as the payments $t_i \not\geq q_i\psi_i$. As a result, for $B \geq 5$, for the negatively correlated scenario, $SMQ$ performs much better.

$PTAS$ instead aims to minimise the average value per unit cost. As it accounts for the actual reserve prices it performs better than $SMQ$ in the negatively correlated case. Additionally, like $SMQ$, it assumes value is additive but also assumes a prior-free environment meaning payments do not need to ensure $t_i \not\geq q_i\psi_i$.

Our proposed mechanisms, $FIN$ and $INF$, perform consistently across budgets and correlations. As we account for the combinatorial nature of the problem and the actual reserve prices the mechanisms provide stable performance. $FIN$ achieves a lower WD, than the other mechanisms, in the uncorrelated and positively correlated scenarios. This is due to the explicit modelling of the aggregation effect, through the 1/n term in the Hoeffding bound. This is particularly pronounced when the budget is higher. $INF$ performs worse than the others expect in the positively correlated case. This is because it underestimates the coalition size effect, however, it is computationally more efficient than $FIN$.

\begin{figure}[htb]
    \centering
    \subfloat[CEN$\left(\rho(W,\theta) = -1\right)$\label{fig:fix_budget_dist_central_n1}]{
    \includegraphics[width=0.3\textwidth]{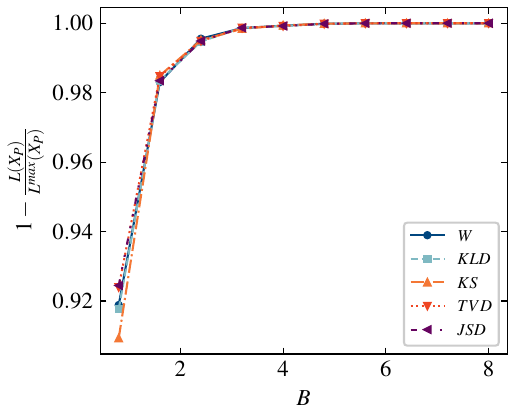}}
    \hfill
    \subfloat[CEN$\left(\rho(W,\theta) = 0\right)$\label{fig:fix_budget_dist_centrals_0}]{
    \includegraphics[width=0.3\textwidth]{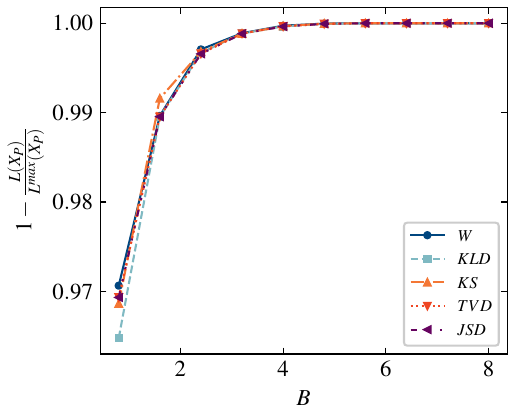}}
    \hfill
    \subfloat[CEN$\left(\rho(W,\theta) = 1\right)$\label{fig:fix_budget_dist_central_1}]{
    \includegraphics[width=0.3\textwidth]{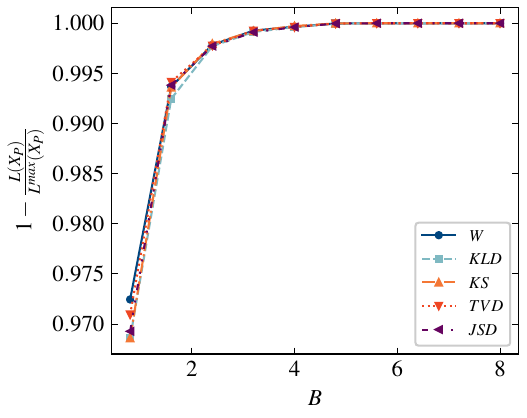}}
    \vfill
    \subfloat[FIN $\left(\rho(W,\theta) = -1\right)$\label{fig:fix_budget_dist_finite_n1}]{
    \includegraphics[width=0.3\textwidth]{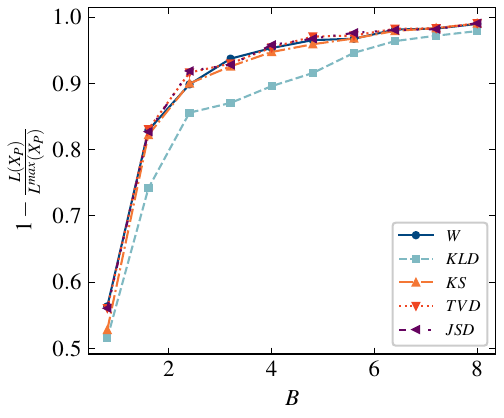}}
    \hfill
    \subfloat[FIN $\left(\rho(W,\theta) = 0\right)$\label{fig:fix_budget_dist_finite_0}]{
    \includegraphics[width=0.3\textwidth]{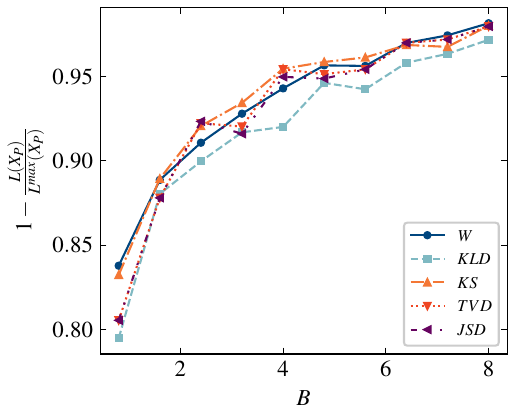}}
    \hfill
    \subfloat[FIN $\left(\rho(W,\theta) = 1\right)$\label{fig:fix_budget_dist_finite_1}]{
    \includegraphics[width=0.3\textwidth]{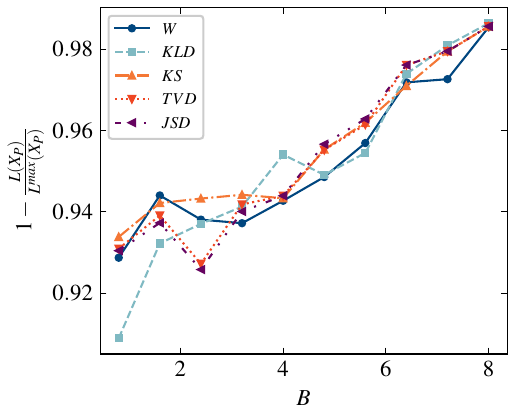}}
    \caption{Accuracy of Exogenous Budget Mechanisms for Mean Estimation using Different Distances}
    \label{fig:fix_budget_dist}
\end{figure}

\paragraph{Statistical Distances}
As discussed in Section \ref{sec:wd_motiv}, we choose the WD as our data valuation metric due to its theoretical properties, however, it is possible to use other distances. Figure \ref{fig:fix_budget_dist} shows the improvement in loss (RMSE) for mean estimation, as a percentage of the worst case loss in the dataset $L^{max}(X_P) = \max_P L(X_P)$, using different distances. In the benchmark case (CEN) we see that performance is very similar across distances. However, using our proposed mechanism, $FIN$, the KLD performs worse than the other distances considered. Overall, performance is consistent across distances suggesting the selection of distance should be based on task-specific or, more generally, on theoretical properties such as those discussed in Section \ref{sec:wd_motiv}.

\paragraph{Unified Valuation Metric} Next, we focus on our main contribution for exogenous budget mechanisms. Namely, we investigate the effect of incorporating both the heterogeneity or non-I.I.D. nature of the data and DP. We compare the performance of the finite formulation for four different scenarios for the WD; (1) DP only $V_i = 1/\epsilon_i$, (2) Non-I.I.D. only $V_i = W(X_i, X_T)$, (3) Exact DP (only for Gaussians as detailed in \cite{Chhachhi2023}) $V_i = W(X_i + X_{DP}, X_T)$ and (4) Upper bound on DP $V_i = W(X_i, X_T) + W(X_{DP},\delta_0)$. Here we also consider the effect correlations, however, in this case we simulate correlations, $\rho(\theta,\epsilon) \in \{-1,0,1\}$, between reserve prices, $\theta$, and the privacy budget, $\epsilon$, rather than the WDs. We assume reserve prices and privacy budgets are both distributed uniformly ($\theta \sim U(0,\bar\theta), \epsilon \sim U(0,\bar\epsilon)$), and DP is achieved using the Gaussian mechanism. To show the significance of our unified metric we consider a budget constrained scenario with $B=0.2\bar\theta N$, a probability of failure $\delta^{dp} = 10^{-15}$, $\bar\theta = 1$, and sweep the upper bound of the uniform distribution of $\epsilon$, $\bar\epsilon \in \{0.1,\dots,100\}$\footnote {We note that the Gaussian mechanism only provides meaningful privacy guarantees when $\epsilon \in (0,1)$, and the probability of failure, $\delta^{dp} \gg 1/N$ \citep{Blanco2023}.}.

\begin{figure}[ht]
    \centering
    \subfloat[$\rho(\epsilon,\theta) = -1$\label{fig:fix_budget_dp_n1}]{
    \includegraphics[width=0.33\columnwidth]{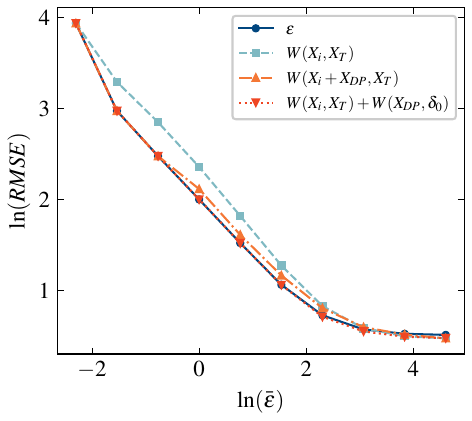}}
    \subfloat[$\rho(\epsilon,\theta) = 0$\label{fig:fix_budget_dp_0}]{
    \includegraphics[width=0.33\columnwidth]{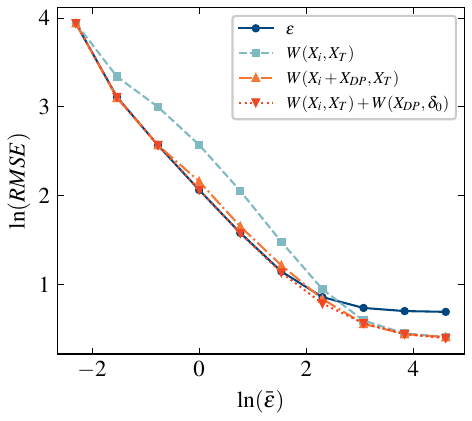}}
    \subfloat[$\rho(\epsilon,\theta) = 1$\label{fig:fix_budget_dp_1}]{
    \includegraphics[width=0.33\columnwidth]{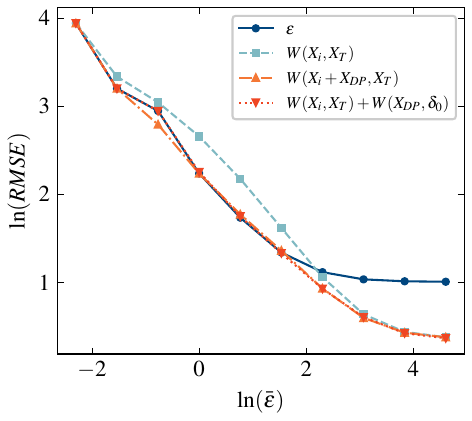}}
    \caption{Effect of DP and Data Heterogeneity}
    \label{fig:fix_budget_dp} 
\end{figure}

Figure \ref{fig:fix_budget_dp} shows the RMSE using the exogenous budget mechanism. When $\epsilon$ is small (high privacy preferences) this is the main driver of value differentiation. As a result, the methods which consider the effect of DP perform better than using only the WD. Conversely, when $\epsilon$ is higher the non-I.I.D. nature is the main driver and methods which include the WD perform better. This difference is more pronounced when $\theta$ and $\epsilon$ are uncorrelated or positively correlated. Both combined approaches perform better across privacy budgets and correlation scenarios.

\subsubsection{Endogenous Budget Mechanisms}\label{sec:proc_endo}
For the endogenous budget and joint optimisation mechanisms we compare against benchmark values. We assume that the buyer is looking to buy data for a specific task and aims to minimise the relevant performance metric/loss function, $L_{\mathcal{M}}(\cdot)$. For example, for median estimation the buyer is aiming to minimise the MAE. We then compare our proposed mechanisms against:
\begin{itemize}
    \item Central Actual ($CEN_\mathcal{M}$): Buyer has access to performance metric for each coalition of data owners and selects the optimal budget feasible coalition\footnote{Minimum $L_{\mathcal{M}}(X_P)-L_{\mathcal{M}}(X_T)$ for the endogenous budget mechanism and minimum $L_{\mathcal{M}}(X_P) - L_{\mathcal{M}}(X_T) + \sum_{i \in P} t_i$ for the joint optimisation mechanism.}. 
    \item Central Distance ($CEN_W$): Buyer has access to the WD for each coalition of data owners and selects the optimal budget feasible coalition\footnote{Minimum $W(X_P,X_T)$ for the endogenous budget mechanism and minimum $K_{\mathcal{M}}W(X_P,X_T) + \sum_{i \in P} t_i$ for the joint optimisation mechanism.}. 
\end{itemize}

We run the mechanisms for a range of loss functions (RMSE, MAE, MPL\textsubscript{$\tau$=0.9,0.8}) and reserve price-distance correlations $\rho(\theta, W) \in \{-1,0,1\}$. We assume the budget provided by the buyer is fixed at $B_{\mathcal{M}}(X_R) = L_{\mathcal{M}}(X_R) - L_{\mathcal{M}}(X_T)$. Instead, we vary the upper bound on the distribution of reserve prices, $\bar\theta \in \{0, 0.2, \dots, 2.4\}$.


\paragraph{Objectives}
\begin{figure}[htb]
    \centering
    \subfloat[Expected Loss ($K_{MAE}\cdot W^{FIN}$)\label{fig:endo_exp_loss}]{
    \includegraphics[width=0.3\textwidth]{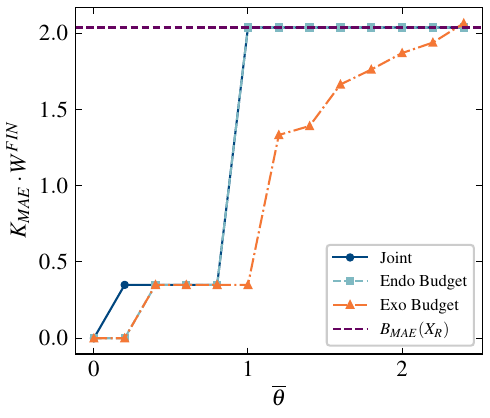}}
    \hfill
    \subfloat[Expected Cost ($\hat\Omega$)\label{fig:endo_exp_cost}]{
    \includegraphics[width=0.3\textwidth]{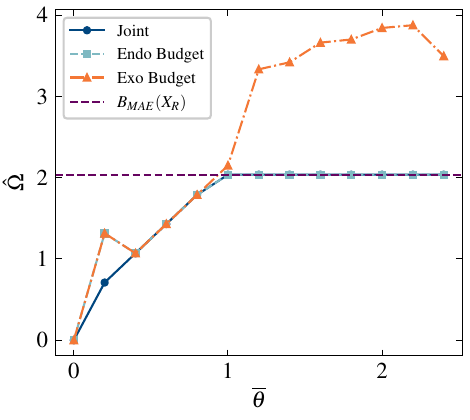}}
    \hfill
    \subfloat[Actual Cost ($\Omega$)\label{fig:endo_act_cost}]{
    \includegraphics[width=0.3\textwidth]{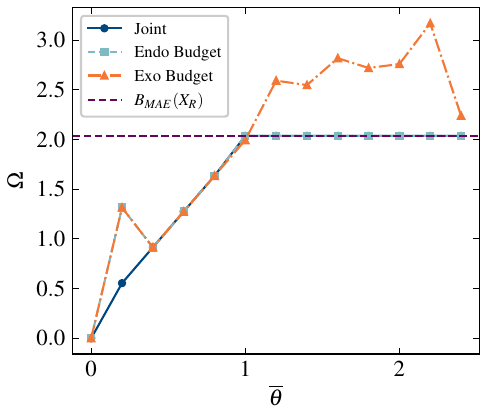}}
    \caption{Median Estimation (MAE) under Different Objectives ($FIN$, $\rho(W,\theta)=-1$)}
    \label{fig:endo_comp}
\end{figure}

Figure \ref{fig:endo_comp} illustrates the dynamics of the proposed mechanisms; exogenous budget, endogenous budget and joint optimisation, for median estimation (minimising the MAE) for a single trial of the finite formulation, $FIN$, assuming the value and reserve prices are negatively correlated. Figure \ref{fig:endo_exp_loss} shows the modelled loss, $K_{\mathcal{M}} \cdot W^{FIN}_P$, determined by the Hoeffding bound, for the procured subset of data, $X_P$. We see that as reserve prices, $\bar\theta$, increases the WD of the procured data increases. For the endogenous budget and joint optimisation mechanisms this does not exceed the reference budget $B_{MAE}(X_R)$. However, for the exogenous budget mechanism the reference budget is exceeded as the decision-dependence is not considered. Next, Figure \ref{fig:endo_exp_cost} and \ref{fig:endo_act_cost}, show the expected cost $\hat\Omega = K_{\mathcal{M}} \cdot W^{FIN}_P + \sum_{i \in P} t_i$ and actual cost $\Omega = (L_{\mathcal{M}}(X_{P}) - L_{\mathcal{M}}(X_T)) + \sum_{i \in P} t_i$, respectively. We see that the exogenous budget mechanism exceeds the reference budget and is therefore not budget feasible in this context. However, the other two mechanisms maintain budget feasibility, even in terms of actual costs as the Hoeffding bound and Lipschitz bound provide an upper bound on the actual loss. The endogenous budget can result in a lower loss, as we see in Figure \ref{fig:endo_exp_loss}, but the overall costs (inc. payments) may be higher. The endogenous budget mechanism is useful in scenarios where the aim is to maximise task performance while maintaining decision-dependent budget feasibility. However, if the buyer also aims to minimise data payments then the joint optimisation approach is most relevant, and will be the focus of the remaining results.

\begin{figure}[htb]
    \centering
    \subfloat[\% Improvement across Benchmarks (MAE)\label{fig:joint_task_n1}]{
    \includegraphics[width=0.3\textwidth]{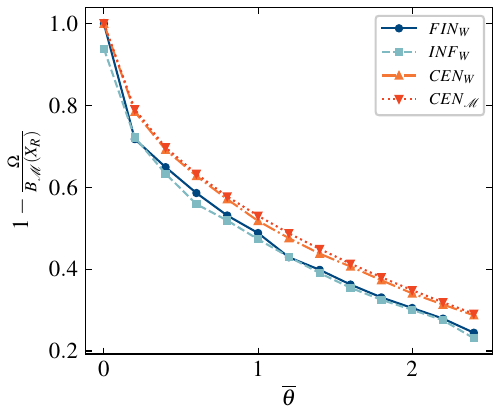}}
    \hfill
    \subfloat[\% Error vs. $CEN_{MAE}$\label{fig:joint_task_0}]{
    \includegraphics[width=0.3\textwidth]{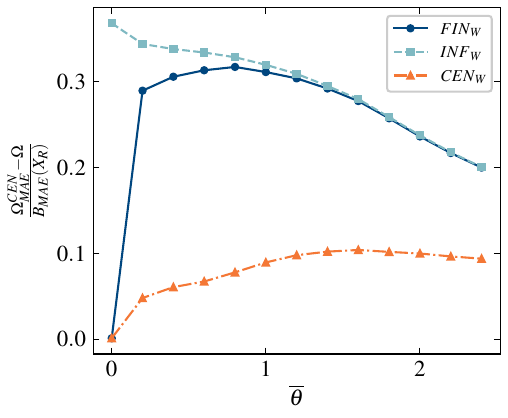}}
    \hfill
    \subfloat[\% Improvement across Tasks\label{fig:joint_task_1}]{
    \includegraphics[width=0.3\textwidth]{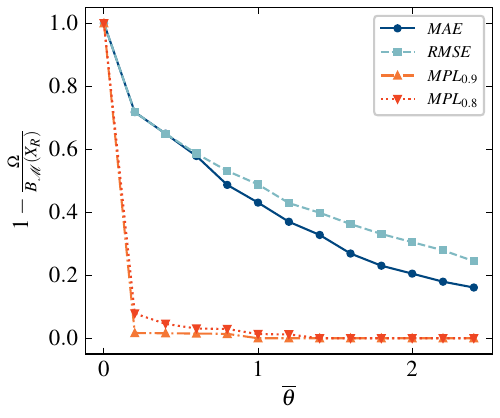}}
    \caption{Joint Optimisation Performance across Tasks and Benchmarks ($\rho(W,\theta) = 0$)}
    \label{fig:joint_task}
\end{figure}

\paragraph{Benchmarks and Tasks}
Having detailed the dynamics of the mechanisms, we now look at the average performance of the joint optimisation mechanism, comparing it against benchmarks and across tasks. Figure \ref{fig:joint_task_n1} shows the percentage improvement in cost compared to the reference for median estimation $\left(1-\frac{\Omega}{B_{MAE}(X_R)}\right)$. We see that the central mechanism using the WD, $CEN_W$, is very similar to the central mechanism using the actual loss values, $CEN_{MAE}$. $FIN_W$ and $INF_W$ perform slightly worse than the central case for median estimation. Figure \ref{fig:joint_task_0} shows the cost difference, in percentage terms, compared to $CEN_{MAE}$, again for median estimation $\left(\frac{\Omega^{CEN}_{MAE}-\Omega}{B_{MAE}(X_R)}\right)$. First, we note that the infinite formulation, $INF_W$, may not select all data sources even when they are free, resulting in a non zero cost difference for $\bar\theta=0$. As the reserve prices increase we see similar cost differences for $FIN_W$ and $INF_W$. We see that the cost difference peaks at around $\bar\theta = 1$ for $INF$ and $FIN$ before decreasing. This is due to the bias introduced by minimising the Hoeffding bound, shown in Figure \ref{fig:hoef_opt_mini}. Indeed, we do not observe this in the central mechanism using the WD, $CEN_W$.

Figure \ref{fig:joint_task_1} shows the average percentage improvement across three different task types; median estimation (MAE), mean estimation (RMSE), quantile estimation (MPL). We see that the finite formulation has similar performance for mean and median estimation but performs badly for quantile estimation (both 90th and 80th). This is due to the tightness of the Lipschitz bound. The MPL, the loss function for quantile estimation, is asymmetric resulting in an overly conservative Lipschitz constant (especially for Gaussian data as seen in Figure \ref{fig:lip_perf}).

\paragraph{Risk Adjustment}
We now investigate the effect of risk by adjusting the confidence level ($\delta \in \{0.1, 0.25, 0.5, 0.75, 0.9,0.95, 0.99\}$), of the Hoeffding bound. One method to tackle the conservativism of the approach is to adjust the confidence level of the Hoeffding bound, $\delta$. By reducing the $\delta$, we are reducing the upper bound on the loss, effectively assuming each data source is more valuable. This can lead to an improvement in the procurement decisions of the mechanism but comes at increasing risk of underestimating the bound. We note that the confidence level indicates that the probability the Hoeffding bound is below the true WD is $1-\delta$. As such, it does not tell us the probability of being below the actual loss, although this is guaranteed to be less than $1-\delta$. 

\begin{figure}[htb]
    \centering
    \subfloat[$\rho(W,\theta)=-1$\label{fig:endo_risk_n1}]{
    \includegraphics[width=0.3\textwidth]{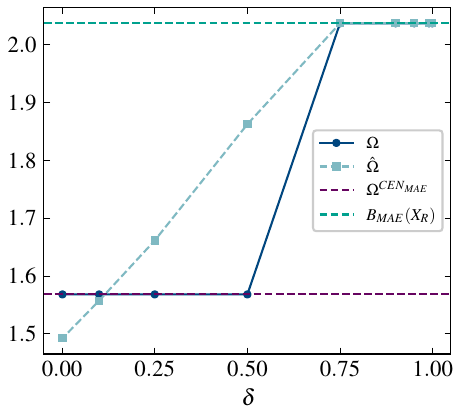}}
    \hfill
    \subfloat[$\rho(W,\theta)=0$\label{fig:endo_risk_0}]{
    \includegraphics[width=0.3\textwidth]{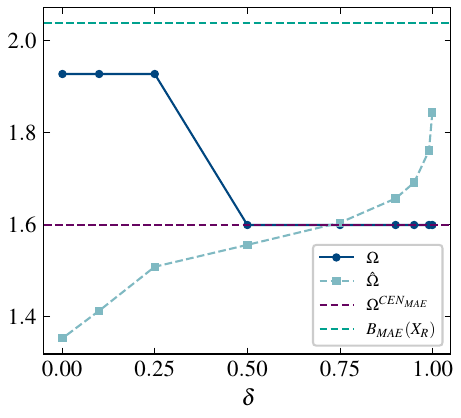}}
    \hfill
    \subfloat[$\rho(W,\theta)=1$\label{fig:endo_risk_1}]{
    \includegraphics[width=0.3\textwidth]{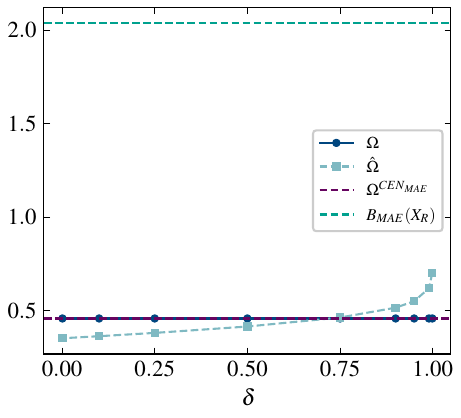}}
    \caption{Risk-Adjustment using $\delta$ adjustment for Median Estimation ($\overline{\theta} = 1.4$)}
    \label{fig:endo_risk}
\end{figure}

Figure \ref{fig:endo_risk} shows how changing $\delta$ affects the modelled and actual procurement costs for median estimation. We plot the actual cost $\Omega$, the modelled cost $\hat\Omega$, the cost achieved in the central case $CEN_{MAE}$ and the reference budget $B_{MAE}(X_R)$.  For $\rho(W,\theta)=-1$, we see that reducing the confidence level still ensures the Lipschitz bound and we are able to achieve the central optimal result when $\delta \leq 0.5$. However, for $\rho(W,\theta)=0$ reducing $\delta$ results in an underestimation of the actual loss. We therefore, end up with increased overall costs. Lastly, when $\rho(W,\theta)=1$, we still underestimate the actual loss when $\delta \leq 0.4$ but this does not lead to a change in the overall cost.

\begin{figure}[htb]
    \centering
    \subfloat[$\rho(W,\theta)=-1$\label{fig:joint_risk_n1}]{
    \includegraphics[width=0.3\textwidth]{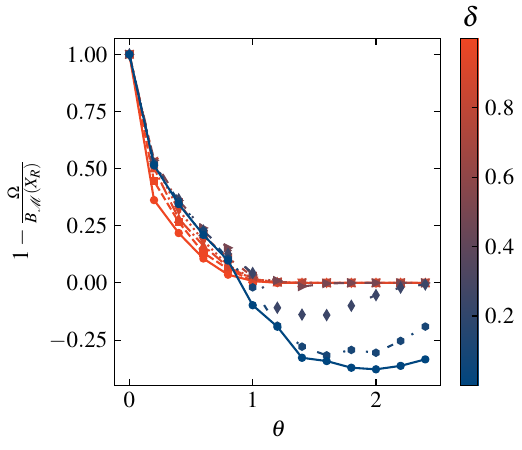}}
    \hfill
    \subfloat[$\rho(W,\theta)=0$\label{fig:joint_risk_0}]{
    \includegraphics[width=0.3\textwidth]{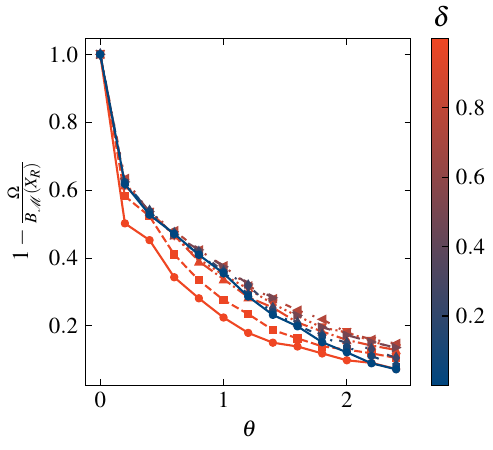}}
    \hfill
    \subfloat[$\rho(W,\theta)=1$\label{fig:joint_risk_1}]{
    \includegraphics[width=0.3\textwidth]{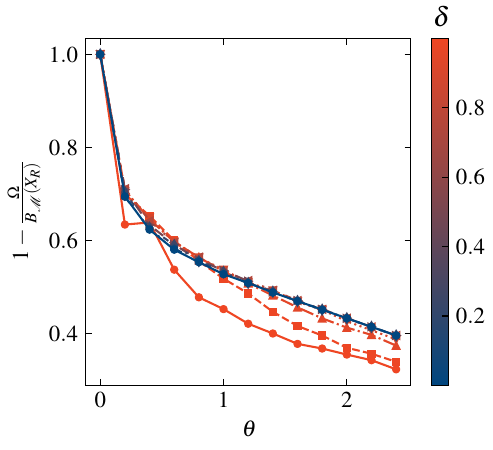}}
    \caption{Average Performance with Risk-Adjustment for Median Estimation ($\overline{\theta} = 1.4$)}
    \label{fig:joint_risk}
\end{figure}

As the effectiveness of the risk adjustment depends on the tightness of the Lipschitz bound as well as correlations between value and reserve prices, we investigate the average effects across 50 trials. Figure \ref{fig:joint_risk} plots the improvement percentage against the reference budget for different values of $\delta$. For the negatively correlated scenario, decreasing $\delta$ results in negative percentages for higher reserve prices. This indicates that in this budget constrained environment, on average, reducing conservatism leads to underestimation of the actual loss, an increase in the overall cost and loss of budget feasibility. Conversely, for the uncorrelated and positively correlated scenarios, reducing conservatism leads to an improvement in overall cost. The correlations are an indication of how much the budget constraints are limiting the selection of valuable data. As such, the negatively correlated scenario, represents the worst-case in this respect and, hence, also results in the highest risk of over procurement.

\paragraph{Levels of Approximation}
The proposed data valuation and procurement mechanisms introduce a number of approximations and bounds to achieve the desired modelling and computational properties. For example, the inclusion of reserve prices to model consumers' WTP/A, the use of the WD instead of the performance metric for a particular task to provide a task-agnostic and privacy-preserving data valuation metric or the Hoeffding bound to avoid the calculation of the WD for each coalition. As such, the mechanism will not perform as well as, for example, a CG mechanism in terms of procurement costs. To understand the levels of approximation, we investigate performance (value of performance metric) under different assumptions:
\begin{itemize}
    \item $Shap$ \citep{Han2020} - Performance achieved, $L(X_T)$, if the buyer had access to all data, and procurement costs are determined using Shapley Values\footnote{The buyer is included as an additional player in the CG, with value being zero in coalitions which do not include the buyer. Implicitly, this assumes data owners' have no reserve prices and therefore no privacy concerns.}.
    \item $CEN_{IR}$ - Assumes a fixed external budget $B_{\mathcal{M}}(X_R)$ and owners have reserve prices. The mechanism selects the coalition, $P$, with minimum, $L(X_P)$, $\text{s.t. } t_i \geq q_i\theta_i, \forall i \in P$.  
    \item $CEN_{IC}$ - Mechanism satisfies IR and IC, that is $t_i \geq q_i\psi_i, \forall i \in P$. 
    \item $CEN_{W}$ - Actual performance replaced by $K \cdot W(X_P,X_T)$.
    \item $CEN_{DP}$ - Include the effect of DP on the WD as in (\ref{eq:ind_wd}).
    \item $FIN$ \& $INF$ - Proposed joint optimisation mechanism where the WDs for each coalition are replaced by the Hoeffding bound.
\end{itemize}
We consider the levels of approximation introduced by our mechanism for three tasks; median, mean and quantile estimation. The cost differences are illustrative, as they vary depending on input values, however, it provides an overview of the effect of the approximations induced and a basis for studying the trade-offs introduced.

\begin{figure}[htb]
    \centering
    \subfloat[Median\label{fig:lvl_approx_mae}]{
    \includegraphics[width=0.3\linewidth]{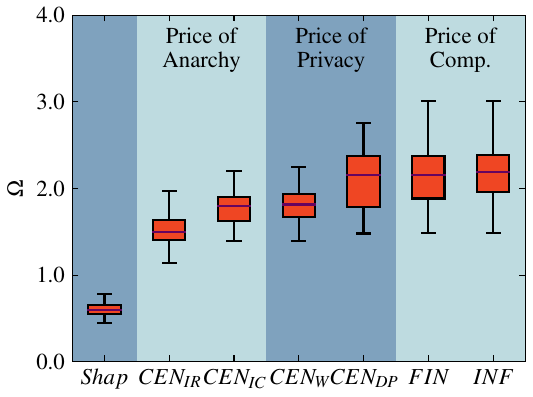}}
    \hfill
    \subfloat[Mean\label{fig:lvl_approx_mse}]{
    \includegraphics[width=0.3\linewidth]{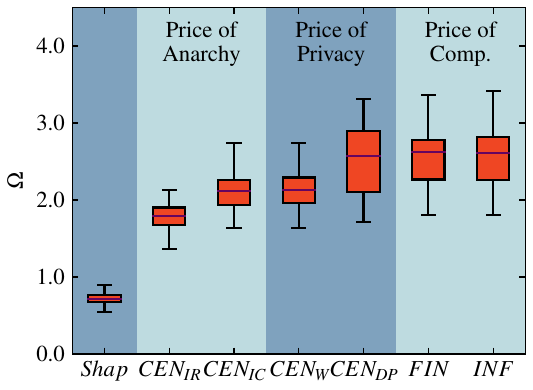}}
    \hfill
    \subfloat[80th Quantile\label{fig:lvl_approx_mpl}]{
    \includegraphics[width=0.3\linewidth]{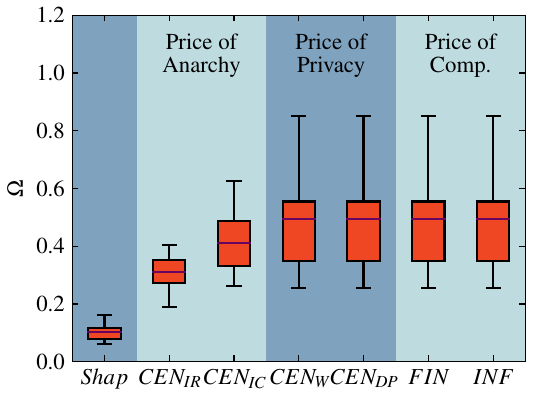}}
    \caption{Waterfall Charts of Costs for Estimating Different Parameters}
    \label{fig:lvl_approx}
\end{figure}

Figure \ref{fig:lvl_approx} shows boxplots of the actual cost ($\Omega$) for each level of approximations under the following conditions; $\bar\theta = 0.8$, $\bar\epsilon = 5$, $\rho(\epsilon,\theta) = 0$ and $\delta = 0.95$.  For all tasks we see that at each level of approximation the mean costs (purple line) increase or saturate. In this example, for mean and median estimation the largest effect is the inclusion of owners' reserve prices ($CEN_{IR}$) and ensuring incentive compatible payments($CEN_{IC}$), or the 'Price of Anarchy' (the cost of selfish behaviour \cite{Bhawalkar2011}). We see that the switch from using actual losses, to using WD ($CEN_{W}$) has a minimal impact on costs, suggesting it is a good valuation metric for mean and median estimation\footnote{Although the MSE is not strictly Lipschitz, we assume a Lipschitz constant, $K_{RMSE} = 1$.}. This, together with the effect of ensuring privacy-preservation of the procured data by adding differentially-private noise, models the price of privacy. Finally, to improve computational tractability, the Hoeffding bounds are used, resulting in the $FIN$ and $INF$ mechanisms. This further increase in costs is the cost of computational efficiency. Interestingly, we see that $FIN$ and $INF$ result in more concentrated costs for mean and quantile estimation, likely due to the reduced set of feasible coalitions these mechanisms can select. For quantile estimation, after introducing incentive compatibility, the costs saturate. Saturation occurs as the approximation results in an overly conservative estimate and the expected costs are higher than the budget $B_{\mathcal{M}}(X_R)$.

\section{Conclusion}\label{sec:conc}
Privacy concerns and a greater understanding of data value among data owners is hindering access to high quality data which is needed to enable data-driven decision making and analytics. Data markets provide a means to value data and compensate data owners for sharing their data thereby providing a means to balance data access and privacy. Existing data market frameworks are unable to adequately model either data owners' (privacy preferences, reserve prices) or buyers' (performance guarantees and the effect of differentially-private noise addition) preferences while simultaneously ensuring privacy-preservation during computationally efficient market clearing. 

In this paper we present a data valuation and procurement mechanism for differentially-private data, based on the WD. We provide a generic framework which ensures strong theoretical performance guarantees for a wide range of tasks/models, allowing for both task-specific and task-agnostic procurement while focusing on ensuring privacy valuation and procurement. We first motivated the use of the WD over other statistical distances before introducing performance guarantees through the Lipschitz bound and endogenously modelling the effect of DP within the WD. To tackle computational issues we provided an approximation scheme using Hoeffding bounds. We then developed three procurement mechanisms, an exogenous budget mechanism for task-agnostic applications, an endogenous budget mechanism for task-specific welfare maximisation, and a joint optimisation mechanism for task performance and data procurement co-optimisation. We derived tractable MISOCP formulations which were extensively tested via simulations with synthetic data distributions. 

The case studies highlighted the suitability of the WD as a valuation metric across a range of tasks, measured through correlations with task specific performance metrics as well as the resulting Shapley allocations. For task-agnostic procurement we showed that our proposed mechanism is more stable than existing mechanisms across budget scenarios and provides a unified metric which is able to capture both the effect of DP on data and the non-I.I.D. nature of data. For task-specific procurement we showed how capturing the decision-dependent structure of data procurement ensures budget feasibility. The implications of approximations/modelling choices that we introduced were investigated in detail. The trade-offs in performance as well as methods to calibrate them through risk adjustment were explored.

Future work will focus on introducing methods to improving the balance between the valuation accuracy, computational tractability, and theoretical guarantees. This includes the development of probabilistic Lipschitz/Hoeffding bounds allowing a more principled manner in which to calibrate budget feasibility and the inherent trade-off between task-specific accuracy and generalisability. In addition, the use of Wasserstein geodesics could be explored as a means to improve combinatorial accuracy. Finally, the mechanism will be extended to a multi-buyer environment to more accurately reflect options available to data owners. 



\acks{We gratefully acknowledge Prof. Pierre Pinson and Dr. Phil Grünewald for their useful comments on earlier versions of this work. This work was supported by the ESRC through the London Interdisciplinary Social Science DTP Studentship (ES/P000703/1:2113082).}



\appendix
\section{Proof of Theorem \ref{lm:lipschitz}}\label{app:proof_lip}
\begin{proof}
    By placing mild assumptions of Lipschitz continuity on the loss function, $l(\cdot)$, we are able to develop a theoretically grounded relationship between the WD between two distributions and the expected difference in the loss function obtained using said distributions.
    \begin{definition}
        (Lipschitz Continuity). 
        Given two metric spaces $(\mathcal{X},d_\mathcal{X})$ where, $d_\mathcal{X}$ denotes a metric on the input set $\mathcal{X}$ and $(\mathcal{Y},d_\mathcal{Y})$ where, $d_\mathcal{Y}$ denotes a metric on the output set $\mathcal{Y}$, a function $l: \mathcal{X} \to \mathcal{Y}$ is Lipschitz continuous if there exists a real constant $K\geq 0$ such that, for all $x_1$ and $x_2$ in $\mathcal{X}$:
        \begin{align}
            d_{\mathcal{Y}}\left(l(x_i),l(x_2)\right) \leq K d_\mathcal{X}(x_1,x_2)
        \end{align}
        where, the smallest such $K$ is known as the Lipschitz constant.
        \label{def:lip_def}
    \end{definition}
    From the definition of Lipschitz continuity, assuming the distance metrics are the $l_1$ norms:
    \begin{align}
        \lvert \mathbb{E}[l(x_1)] - \mathbb{E}[l(x_2)]\rvert \leq \lvert \mathbb{E}[x_1] - \mathbb{E}[x_2]\rvert         \label{eq:lip}
    \end{align}
    The dual formulation of the WD, (\ref{eq:wass_dual}), is an upper bound on the rhs of (\ref{eq:lip}).
\end{proof}
\section{Proof of Theorem \ref{th:hoeffding}}\label{app:proof_hoef}
\begin{proof}
    The triangle inequality bounds the WD of $X_P$:
    \begin{align*}
        W\left(\frac{1}{|P|}\sum_{i\in P}X_i,X_T\right) \leq \frac{1}{|P|}\sum_{i\in P} W(X_i,X_T)
    \end{align*}
    We view $W_P$ as a bounded random variable on the interval $ \left[ 0, \frac{1}{|P|}\sum_{i\in P} W(X_i,X_T)\right]$. 
    Applying the Hoeffding inequality and noting that $\mathbb{E}\left[W(\sum_{i=1}^{N}X_i, X_T)\right]=0$, we obtain:
    \begin{align}
    \begin{split}
        P\left\{W \left( \frac{1}{|P|} \right. \right. \left. \left. \sum_{i \in P}X_i, X_T\right) \geq t \right\} \leq  2\exp\left(\frac{-2|P|^2t^2}{\sum_{i\in P}W\left(X_i, X_T\right)^2}\right)
    \end{split}
    \end{align}
    For a specified confidence level, $\delta \in [0,1)$, we get:
    \begin{align}
        t_{\delta}(P) \leq \sqrt{
    \frac{\sum_{i\in P}W\left(X_i, X_T\right)^2 \ln\left( \frac{2}{1-\delta}\right)}{2|P|^2}}
    \label{eq:hoef_inf}
    \end{align}
    Finally, to account for the finite sample $N$, we introduce a finite sample correction factor of $\left(\frac{N-|P|}{N}\right)$  \citep{Yan2014}.
\end{proof}

\section{Proof of Monotonicity of (\ref{opt:fin_misocp})}\label{app:mono}
\begin{proof}
The reformulation in (\ref{opt:fin_misocp}) is exact, allowing us to analyse (\ref{opt:fin_or}) directly \citep{Fallah2022}. Let $q = [q_0,\dots, q_N]$ be the optimal solution of (\ref{opt:fin_or}) for $\theta = [\theta_1,\dots, \theta_N]$. Now, suppose we increase, without loss of generality, $\theta_1$ such that $\theta_1'>\theta_1$ and $\theta_i' = \theta_i \enspace\forall i >1$. Then $q' = [q'_0,\dots, q'_N]$ is the resulting optimal solution of (\ref{opt:fin_or}). If the true reserve price is $\theta$:
\begin{align}
\begin{split}
        &q_0 B(X_R) +  C^{FIN}g(q) + \sum\limits_{i \in \mathcal{N}} q_i \psi_i(\theta_i) \leq q_0' B(X_R) + C^{FIN}g(q') + \sum_{i \in \mathcal{N}} q_i' \psi_i(\theta_i)
\end{split}
\end{align}
Similarly, if the true reserve price vector is $\theta'$:
\begin{align}
\begin{split}
    & q_0' B(X_R) + C^{FIN}g(q') + \sum_{i \in \mathcal{N}} q_i' \psi_i(\theta_i') \leq q_0 B(X_R) + C^{FIN}g(q) +  \sum_{i \in \mathcal{N}} q_i \psi_i(\theta_i')
\end{split}
\end{align}
where, $g(x) = \sqrt{\sfrac{\left(N-\sum_{i \in \mathcal{N}} x_i\right)\sum^{N}_{i=1} x_iW_i^2}{\left(\sum_{i \in \mathcal{N}} x_i\right)^2}}$. 

Taking the summation of both sides of the inequalities and given that $\theta_i' = \theta_i \enspace\forall i >1$, we obtain:
\begin{align}
    \left(q_1 - q_1'\right)\left(\psi_1(\theta_1) - \psi_1(\theta_1')\right) \leq 0
\end{align}

\begin{assumption}[Regular Distribution]
    The reserve price distribution is regular, $\psi_i(\theta)$ is increasing in $\theta$,  $ \forall i \in N$.
\label{ass:regular}
\end{assumption}
As discussed in \cite{Fallah2022}, Assumption \ref{ass:regular} is standard in mechanism design, in particular for procurement auctions such as ours. Distributions which have this property are also called regular distributions, and include Gaussians, uniform and exponential distributions. Assumption \ref{ass:regular} and the above inequality show that the point-wise optimisation problem (\ref{opt:fin_or}) is monotonically decreasing in $\theta$ ($\left(q_1 - q_1'\right) \geq 0$).
\end{proof}

\section{MISOCP Formulation under Infinite Population Assumption}\label{app:inf_misocp}

\begin{subequations}
    \begin{align}
        \min_{q,s,z} \quad & q_0 B_{\mathcal{M}}(X_R) +  C^{INF}s + \sum^{N}_{i \in \mathcal{N}}q_i \psi_i\tag{\theparentequation}\\
        \text{s.t.} \quad & \lVert q_iW_i \rVert \leq \sum^{N}_{i \in \mathcal{N}} z_i \\
        & 0 \leq z_i \leq  Mq_i, \quad \forall i \in \mathcal{N}\\
        & 0 \leq s - z_i \leq  M(1-q_i), \quad \forall i \in \mathcal{N}\\
        & 1 \leq \sum^{N}_{i=0} q_i  \leq N\\
        &q \in \{0,1\}^{N+1} , s \in \mathbb{R}_{+}, z \in \mathbb{R}_{+}^N
    \end{align}
    \label{opt:inf_misocp}
\end{subequations}
where, $C^{INF} =  K_{\mathcal{M}}\sqrt{\frac{\ln\left(\frac{2}{1-\delta}\right)}{2}}$ and $M > \lVert W\rVert$.
\bibliography{references,add_refs}

\end{document}